\documentclass[10pt,twocolumn,letterpaper]{article}

\usepackage[pagenumbers]{wacv} % To force page numbers, e.g. for an arXiv version

\usepackage{graphicx}
\usepackage{amsmath}
\usepackage{amssymb}
\usepackage{booktabs}

\usepackage{multirow} % For multirow cells
\usepackage{array}    % For \newcolumntype

\usepackage[pagebackref,breaklinks,colorlinks]{hyperref}

\usepackage[capitalize]{cleveref}
\crefname{section}{Sec.}{Secs.}
\Crefname{section}{Section}{Sections}
\Crefname{table}{Table}{Tables}
\crefname{table}{Tab.}{Tabs.}

\def\wacvPaperID{1399} % *** Enter the WACV Paper ID here
\def\confName{WACV}
\def\confYear{2025}

\usepackage{arydshln}
\usepackage{multirow}
\usepackage{amsmath}
\usepackage{comment}
\usepackage{algorithm}
\usepackage{algpseudocode}
\usepackage{orcidlink}
\usepackage{color,soul}
\usepackage{xcolor}
\definecolor{cerulean}{rgb}{0.0,0.48,0.65}
\definecolor{green}{rgb}{0.01, 0.75, 0.24}
\definecolor{Black}{RGB}{0.0, 0.0, 0.0}
\newcommand{\red}[1]{\textcolor{red}{#1}}
\newcommand{\brown}[1]{\textcolor{black}{#1}}
\newcommand{\blue}[1]{\textcolor{Black}{#1}}

\newcommand{\br}[1]{\textcolor{Black}{#1}}

\newcommand{\teal}[1]{\textcolor{Black}{#1}}
\newcommand{\shadow}[1]{}
\def\r{\red}
\def\b{\blue}

\def\s{\shadow}

\def\t{\teal}

\begin{document}

\title{Graph-Jigsaw Conditioned Diffusion Model for Skeleton-based Video Anomaly Detection}

\author{
Ali Karami$^{1,2}$, Thi Kieu Khanh Ho$^{1,2}$, Narges Armanfard$^{1,2}$\\
$^1$Department of Electrical and Computer Engineering, McGill University\\
$^2$Mila - Quebec AI Institute, Montreal, QC, Canada\\
{\tt\small \{ali.karami, thi.k.ho, narges.armanfard\}@mail.mcgill.ca}
}

\maketitle

%%%%%%%%% ABSTRACT
\begin{abstract}
  Skeleton-based video anomaly detection (SVAD) is a crucial task in computer vision. Accurately identifying abnormal patterns or events enables operators to promptly detect suspicious activities, thereby enhancing safety. Achieving this demands a comprehensive understanding of human motions, both at body and region levels, \brown{while also accounting for the wide variations of performing a single action}. However, existing studies fail to simultaneously address these crucial properties. This paper introduces a novel, practical, and lightweight framework, namely \underline{\textbf{G}}raph-J\underline{\textbf{i}}gsaw \underline{\textbf{C}}onditioned D\underline{\textbf{i}}ffusion Model for \underline{\textbf{S}}keleton-based Video \underline{\textbf{A}}nomaly \underline{\textbf{D}}etection (GiCiSAD)  to overcome the challenges associated with SVAD. GiCiSAD consists of three novel modules: the Graph Attention-based Forecasting module to capture the spatio-temporal dependencies inherent in the data, the Graph-level Jigsaw Puzzle Maker module to distinguish subtle region-level discrepancies between normal and abnormal motions, and the Graph-based Conditional Diffusion model to generate a \t{wide} spectrum of human motions. Extensive experiments on four widely used skeleton-based video datasets show that GiCiSAD outperforms existing methods with significantly fewer training parameters, establishing it as the new state-of-the-art.
\end{abstract}

\section{Introduction}
\label{sec:intro}

Skeleton-based video anomaly detection (SVAD) is an important task in computer vision and video surveillance \cite{morais2019learning,markovitz2020graph,luo2021normal,flaborea2023contracting,flaborea2023multimodal,xiao2023human,pang2024detecting}. It refers to a task of identifying abnormal behaviors or motions that deviate from the typical patterns observed in normal activities. Unlike conventional video anomaly detection (VAD) \cite{sultani2018real,zhao2017spatio,nguyen2019anomaly,ramachandra2020survey,wang2022video}, SVAD involves skeleton-based representations, which focus on the key joints and their connections, effectively capturing more concise and essential information of human activities in video sequences, while reducing the computational complexity compared to the pixel-level analysis. However, SVAD datasets pose several critical challenges for anomaly detection algorithms. 

First, skeleton-based video data is inherently a time-series data, which exhibits \textit{spatio-temporal dependencies} \cite{ho2023graph,nikpour2023spatio}. In essence, spatial dependencies signify the relationships among skeleton joints within a frame such as body posture, gestures, and interactions. Meanwhile, temporal dependencies are represented by the temporal evolution of skeletal motions that capture the dynamics of human activities over time.
Understanding these spatio-temporal dependencies is crucial for distinguishing normal and abnormal motions. For example, deviations from expected spatial arrangements or sudden changes in joint trajectories over time may indicate potential anomalies. Therefore, by analyzing the spatio-temporal evolution of skeletal joints, anomaly detection algorithms can attain a semantic understanding of human activities. Recently, graph-based approaches \cite{deng2021graph,kim2022graph,ren2023graph,ho2023graph,ho2023self,hojjati2023multivariate,ho2023multivariate} have gained significant attention in time-series data due to their capabilities of dynamically learning graphs to effectively capture both types of dependencies, making them well-suited for SVAD tasks. \br{Although several graph-based studies have been conducted for SVAD  \cite{markovitz2020graph,luo2021normal,li2022human}, none of them have dynamically learned the evolving relationships between joints, which are essential for capturing the dynamic nature of human activities.}

% Understanding these spatial relationships is important for distinguishing between normal and abnormal motion patterns. For example, certain joint configurations denote specific actions, while deviations from expected spatial arrangements are deemed as anomalies. 

% \r{Understanding how these changes over time is crucial to distinguish normal actions from abnormal events. For instance, sudden changes in joint trajectories may indicate potential anomalies.} 

\br{Second, subtle differences between normal and abnormal actions can oftentimes be localized to specific regions of the body rather than affecting the entire body.} \b{This is while all existing SVAD methods are based on modeling the human body as a whole and ignore the importance of such local variations when detecting anomalies \cite{rodrigues2020multi,jain2021posecvae,flaborea2023multimodal,stergiou2024holistic}. In the presence of a localized anomaly, these holistic-based models tend to classify the activity as normal since the majority of the body regions are acting normally except for a small region.} For instance, consider a scenario where a person is walking normally, but their arm exhibits abnormal movements due to injury. We refer to this issue as \textit{region-specific discrepancies}. \t{Recently, self-supervised learning (SSL) \cite{liu2021self,li2021cutpaste,zou2022spot,bozorgtabar2023attention,ristea2022self,hojjati2024self} has emerged as a promising research direction for VAD. Unlike unsupervised methods, which learn directly from unlabeled data to identify patterns indicative of anomalies, SSL goes a step further by defining pretext tasks that encourage the model to focus on region-level features \cite{wang2022video}. While SSL has been widely used in the context of the image domain \cite{misra2020self,taleb2021multimodal,taleb20203d,bucci2021self}, it remains unanswered how to adapt this approach to the field of SVAD, particularly considering the presence of skeleton data instead of traditional images in this context.}

% \br{Second, subtle differences between normal and abnormal actions can oftentimes be localized to specific regions of the body rather than affecting the entire body. \b{This is while all existing SVAD methods are based on modeling the human body as a whole and ignore the importance of such local variations \r{[CITE]} when detecting anomalies. In the presence of a localized anomaly, these holistic-based models tend to classify the activity as normal, since the majority of the body regions are acting normally except for a small region, such as an arm. We refer to} this issue as \textit{region-specific discrepancies}.}

\s{In such cases, \b{detection} based solely on \b{a global model that has learned the overall body} spatio-temporal dependencies can be ineffective. For instance, consider a scenario where a subject is walking normally, but their arm exhibits abnormal movements due to injury. In this case, relying solely on the overall body structure \b{as a whole} might fail to detect the anomaly, as the majority of the body's motion remains normal \b{and the small anomalies happening in sub-body parts can not be captured by a global model}.
We refer this issue as \textit{region-specific discrepancies}.}

\br{Third, when dealing with skeleton-based video data, it is essential to acknowledge that there are \textit{infinite variations} of performing both normal and abnormal actions \cite{nayak2021comprehensive,flaborea2023multimodal}. In other words, both normal and abnormal behavior can be complex and multifaceted, encompassing a wide range of actions, gestures, and interactions. While some studies \cite{flaborea2023contracting,luo2021normal,markovitz2020graph,morais2019learning} focused on generating a single reconstruction of the input data, these approaches often fail to capture the wide spectrum of human motions. Moreover, while recent research has addressed the diversity of both normal and abnormal activities \cite{flaborea2023multimodal}, by considering the body as a whole, they overlook the fact that abnormalities may be localized to only specific regions of the body, potentially leading to misdetection in cases where anomalies occur in isolated regions while the rest of the body remains normal.\s{\r{THERE IS A STRONG FOCUS ON such region-specific anomalies, do you have any proof that the method can detect them but others miss to detect... any visualization, metric...}}} \footnote{\brown{We present the Related Works section in Section A of the Supplementary Material.}}

\br{Recognizing the critical importance of spatio-temporal dependencies, region-specific discrepancies, and infinite variations inherent in skeleton-based video data, in this paper, we propose \underline{\textbf{G}}raph-J\underline{\textbf{i}}gsaw \underline{\textbf{C}}onditioned D\underline{\textbf{i}}ffusion Model for \underline{\textbf{S}}keleton-based Video \underline{\textbf{A}}nomaly \underline{\textbf{D}}etection - hereafter GiCiSAD. Essentially, GiCiSAD includes three novel modules. The Graph Attention-based Forecasting module leverages a graph learning strategy to effectively capture the spatio-temporal dependencies. To address the issue of region-specific discrepancies, we propose a novel graph-level SSL with a difficult pretext task called Jigsaw puzzles \cite{wang2022video,markaki2023jigsaw}. We name this module Graph-level Jigsaw Puzzle Maker, which involves various subgraph augmentations applied to the learnable graph, hence providing supervisory signals to help GiCiSAD capture a slight region-level difference between normal and abnormal behaviors. Lastly, to address the infinite variations, GiCiSAD incorporates a newly proposed diffusion-based model called the Graph-level Conditional Diffusion Model, which utilizes the learned graph from past frames as conditional information to generate diverse future samples.}

 % \r{It is hypothesized that, in the inference phase, by ensuring consistency among generated normal samples compared to true normal samples and inconsistency among generated abnormal samples compared to the abnormal ground truths, GaLaSAD can effectively differentiate between normal and abnormal behavior. }

% In summary, the main contributions of this paper are described as follows:
In summary, our contributions are as follows:

\begin{itemize}
    \item  \br{The first study in the SVAD field that \b{presents a unified framework for} effective tackling of the\s{comprehensive} challenges posed by the spatio-temporal dependencies, region-specific discrepancies, and infinite variations inherent in skeleton-based video data.}
    %
    % \item A novel ensemble network, incorporating three modules: Graph Attenion-based Forecasting to dynamically learns dependencies between body joints across spatial and temporal dimensions, Graph-level Jigsaw Puzzle Maker aimed at achieving more discriminative region-level understanding in SVAD, and  Graph-level Conditional Diffusion Model to facilitate generating diverse future motion patterns, leveraging past motions as a guide. 
    % %
    % \item \br{The first graph self-supervised learning study in the SVAD field aimed at achieving more discriminative region-level understanding in SVAD. \r{revise and emphasize more on this}}
    
    \item A novel graph attention-based approach dynamically learns dependencies between body joints across spatial and temporal dimensions. 
    % This method is adaptable and not limited to a specific encoder or a pre-defined number of joints.
%     \br{A novel graph attention-based approach that dynamically learns the dependencies between body joints across both spatial and temporal dimensions. (not bounded to a specific encoder with pre-defined number of joints as the task can be adapted regardless of the joint's
% settings.)}
    %
    \item \br{The first graph self-supervised learning study aimed at achieving more discriminative region-level understanding in SVAD.}
    \item \br{A novel graph-level conditional diffusion model to facilitate generating diverse future motion patterns, leveraging past motions as a guide.}
    \item A thorough validation on four widely used SVAD datasets showcases our superior anomaly detection performance and a remarkable 40\% reduction in training parameters compared to state-of-the-art (SOTA).
    % \item Extensive experiments conducted on four widely used skeleton-based video datasets conclusively demonstrate GiCiSAD's superior performance over existing methods. Moreover, GiCiSAD poses a significantly reduced number of training parameters up to 40\% compared to existing methods. This solidifies our method's position as the new state-of-the-art (SOTA) in the field.
\end{itemize}

\section{Proposed Method} \label{sec:proposed_method}

We define a skeleton-based video dataset that consists of human poses over time as $X=\{\mathbf{x}_{(i)}\}_{i=1}^N$, where $\mathbf{x}_{(i)} = (x_{(i)}^1,x_{(i)}^2,\ldots,x_{(i)}^K)$ is the $i$th observation in the frame sequences of $N$ observations, $\mathbf{x}_{(i)} \in \mathbb{R}^{K \times L}$. $K$ and $L$, respectively, denote the number of joints and the number of frames in the $i$th observation. An observation can be conceptualized as a sliding window of size $L$. We further divide $\mathbf{x}_{(i)}$ into two parts: the first (past) frames  $\mathbf{x}_{(i)}^{1:l}$ and the latter (future) frames  $\mathbf{x}_{(i)}^{l+1:L}$, \br{where $l$ regulates the size of past and future frames}. For simplicity, we present $\mathbf{x}_{(i)}^{1:l}$ and $\mathbf{x}_{(i)}^{l+1:L}$ as $\mathbf{x}^{-}$ and $\mathbf{x}^{+}$, respectively. Our task is to detect the frames with abnormal poses in the test data by training the model with only normal data. 

The block diagram of the proposed GiCiSAD method, depicting the three novel modules, namely Graph Attention-based Forecasting, Graph-level Jigsaw Puzzle Maker, and Graph-based Conditional Diffusion Model, is shown in \cref{fig:model}. Details of each module are presented below. For the sake of completeness, we provide the pseudocode in Section B of the Supplementary Material. 

% \begin{figure}[tb]
%   \centering
%   \includegraphics[width=0.9\textwidth]{images/model.png} 
%   \caption{\br{The overall framework of GiCiSAD.}}
%   \label{fig:model}
% \end{figure}

\begin{figure*}[t]
  \centering
  \includegraphics[width=0.7\textwidth]{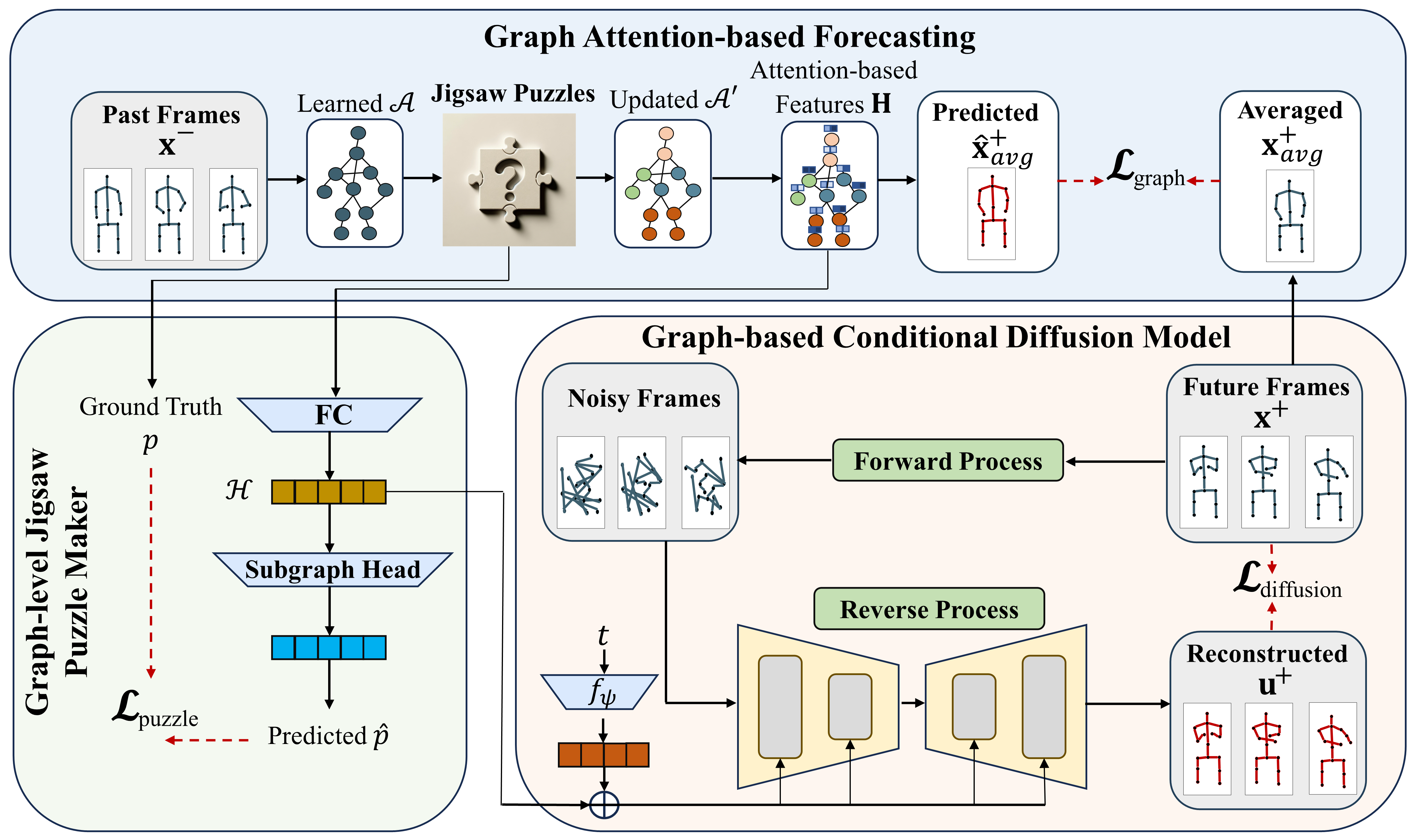}

   \caption{The overall framework of GiCiSAD.}
   \label{fig:model}
\end{figure*}

\subsection{Graph Attention-Based Forecasting} \label{sec:graph_attention}
In this section, our objective is to capture both the spatial dependencies between joints within each frame and the temporal dependencies across frames. \b{This is to provide a model that is more sensitive to capture the holistic structure of the body.} We achieve this by constructing a graph for the past frames, i.e.,  $\mathbf{x}^{-}$, and performing the task of forecasting future frames, i.e., $\mathbf{x}^{+}$, based on the past. 
We first represent $\mathbf{x}^{-}$, as a graph denoted as $\mathcal{G}= \{\mathbf{H}, \mathcal{A}\}$, where $\mathbf{H} \in \mathbb{R}^{D \times K}$ is the representation matrix of all $K$ nodes (joints), $D$ is a hyperparameter defining the feature dimension, and $\mathcal{A} \in \mathbb{R}^{K \times K}$ is the adjacency matrix that encodes the relationship between joints. We learn $\mathbf{H}$ and $\mathcal{A}$ as below.

Starting with $\mathcal{A}$, in the context of human motion analysis, different joints have different characteristics, and these characteristics can be related in complex ways. It is crucial to represent each joint in a flexible way that captures the different underlying patterns. We do this by introducing a feature vector for each joint. For instance, the feature vector for the $k$th joint is denoted as $\mathbf{v}_k \in \mathbb{R}^D$, $k \in \{1,2,\ldots,K\}$. Note that these feature vectors are initialized randomly and then trained along with the rest of the model. To this end, we construct $\mathcal{A}$, where each element $\mathcal{A}_{kn}$ represents the relationship between the feature vectors $\mathbf{v}_{k}$ and $\mathbf{v}_{n}$ of the $k$th joint and the $n$th joint, respectively. To learn $\mathcal{A}$, we first compute the cosine similarity \cite{deng2021graph} between the $k$th node's feature vector and the feature vectors of its candidates named $\mathcal{C}_k$, i.e., all $K$ nodes, excluding $k$ as below:
\begin{equation} \label{eq:cosine_sim}
\text{Sim}_{kn} = \frac{\mathbf{v}_k^{\top} \cdot \mathbf{v}_n}{\|\mathbf{v}_k\| \cdot \|\mathbf{v}_n\|} \quad \text{for } n \in \mathcal{C}_k
\end{equation}

Then, for each node $k$, we select the Top$\delta$ highest values of cosine similarities in its candidates and consider that the $k$th node is connected to those nodes. Hence, each node has $\delta$ connections. $\delta$ is a hyperparameter defining the sparsity level based on specific applications at hand. This process is shown as below: 

\begin{equation}
\label{eq:adj_matrix}
\mathcal{A}_{kn} = \begin{cases} 
1 & \text{if } n \in \text{Top}\delta(\{\text{Sim}_{kn'} : n' \in \mathcal{C}_k\}) \\
0 & \text{otherwise}
\end{cases}
\end{equation}

Note that \br{the connection from node $k$ to node $n$ indicates that the feature vector of node $k$ is used for modeling the behavior of node $n$}. It is worth mentioning that we use a \textit{directed} graph since the dependency patterns between nodes are not necessarily symmetric. 

Next, we learn $\mathbf{H}$ by a graph-based attention mechanism \cite{deng2021graph}, which leverages $\mathcal{A}$. However, $\mathcal{A}$ is first fed into the Graph-level Jigsaw Puzzle Maker module, and the output of this module is a permuted version of the adjacency matrix, namely $\mathcal{A}'$. This modified $\mathcal{A}'$ serves as the input for the graph-based attention mechanism. Further details regarding the Graph-level Jigsaw Puzzle Maker module is provided in \cref{sec:puzzel_maker}. Essentially, the graph-based attention mechanism is to integrate a node’s information with that of its neighbors, guided by the learned and permuted graph structure, i.e., $\mathcal{A}'$. We define the attention coefficient between two nodes $k$ and $n$ as $\alpha_{kn}$ as below:

\begin{equation}
\alpha_{kn} = \frac{\exp\left(\text{LeakyReLU}\left(\mathbf{s}^\top(\textbf{g}_k \oplus \textbf{g}_n)\right)\right)}{\sum_{j \in \mathcal{N}(k) \cup \{k\}} \exp\left(\text{LeakyReLU}\left(\mathbf{s}^\top(\mathbf{g}_k \oplus \mathbf{g}_j)\right)\right)},
\end{equation}
where $\mathbf{g}_k = \mathbf{v}_k \oplus \mathbf{W}\mathbf{x}_{k}^{-}$, $\mathbf{W} \in \mathbb{R}^{D \times l}$ is a trainable weight matrix applied to every node, $\mathbf{x}_{k}^{-} \in \mathbb{R}^{l}$ is the $k$th node's input value over the past frames, $\oplus$ denotes concatenation, $\mathcal{N}(k) = \{n | \mathcal{A'}_{kn} > 0\}$ is the set of neighbors of the node $k$ obtained from $\mathcal{A'}$, and \br{the vector of learned coefficients is denoted by $\mathbf{s}$.}  

Then, we obtain the representation vector for the $k$th node as below:

\begin{equation}
\mathbf{h}_k = \text{ReLU}\Big(\alpha_{k,k}\mathbf{W}\mathbf{x}_k^{-} + \sum_{n \in \mathcal{N}(k)} \alpha_{k,n}\mathbf{W}\mathbf{x}_n^{-}\Big).
\end{equation}

We then element-wise multiply (denoted as $\odot$) the representation vector of each node, i.e., $\mathbf{h}_k$, with its corresponding feature vector, i.e., $\mathbf{v}_k$, and the output will be fed into stacked fully-connected layers, \br{i.e., $f(\theta)$}, with the output dimension of $K$ to predict an average of future frames, denoted as $\hat{\mathbf{x}}^{+}_{\text{avg}}$. \brown{ We observed that computing the graph loss on the average of temporal frames or per-frame yields comparable performance. We chose the average approach as it benefits both speed and smoothness of convergence. Note that the graph loss serves as additional supervision to facilitate the training of the conditioning network, essential as the training of the diffusion model relies on a well-trained conditioning.} $\hat{\mathbf{x}}^{+}_{\text{avg}}$ is calculated as below:

% \r{WHY AVERAGE?, sth like we observe that predicting average leads to a more smooth and easier convergence, while achieving better performance .. Reason for this... we provide a comparison of predicting average and individual frames in Appendix-Section..} 

\begin{equation}
\hat{\mathbf{x}}^{+}_{\text{avg}} = f_\theta \big( \mathbf{v}_1 \odot \mathbf{h}_1, \ldots, \mathbf{v}_K \odot \mathbf{h}_K \big) 
\end{equation}

% \r{Despite that the algorithm implicitly looking at the subgraph because of $\delta$, but the loss is holistic. This mean subtle changes get smeared in the majority of the parts that are normal. hidden }

\t{We aim to capture the holistic structure of the body by optimizing $\mathcal{L}_\text{graph}$ denoted in \cref{eq:loss_gaf}, which involves utilizing} the Mean Square Error between the predicted $\hat{\mathbf{x}}^{+}_{\text{avg}}$ and the actual average of future frames, denoted as $\mathbf{x}^{+}_{\text{avg}}$.

% We use the Mean Square Error between the predicted $\hat{\mathbf{x}}^{+}_{\text{avg}}$ and the actual average of future frames, i.e., $\mathbf{x}^{+}_{\text{avg}}$, as the loss of Graph Attention-based Forecasting module:

\begin{equation} \label{eq:loss_gaf}
\mathcal{L}_{\text{graph}} = \left\| \hat{\mathbf{x}}^{+}_{\text{avg}} - \mathbf{x}^{+}_{\text{avg}} \right\|^2_2
\end{equation}

The representation matrix $\mathbf{H}$ of the past frames is the combination of the representation vectors of all nodes as $\mathbf{H}= \{\mathbf{h}_1,\mathbf{h}_2, \ldots, \mathbf{h}_K\}$. We then project $\mathbf{H}$ to a fully connected layer (FC) and output $\mathcal{H} \in \mathbb{R}^{D}$, which is used to solve the task of the Graph-level Jigsaw Puzzle Maker module described in the next section. Simultaneously, $\mathcal{H}$ is also used as the conditioning signal for the Graph-based Conditional Diffusion Model, described in \cref{sec:diffusion}. 

\subsection{Graph-level Jigsaw Puzzle Maker} \label{sec:puzzel_maker}

\s{\r{you may put the SSL part of the intro here and revise the below paragraph accordingly.}}

\begin{figure}[t]
  \centering
  \includegraphics[width=1.02\linewidth]{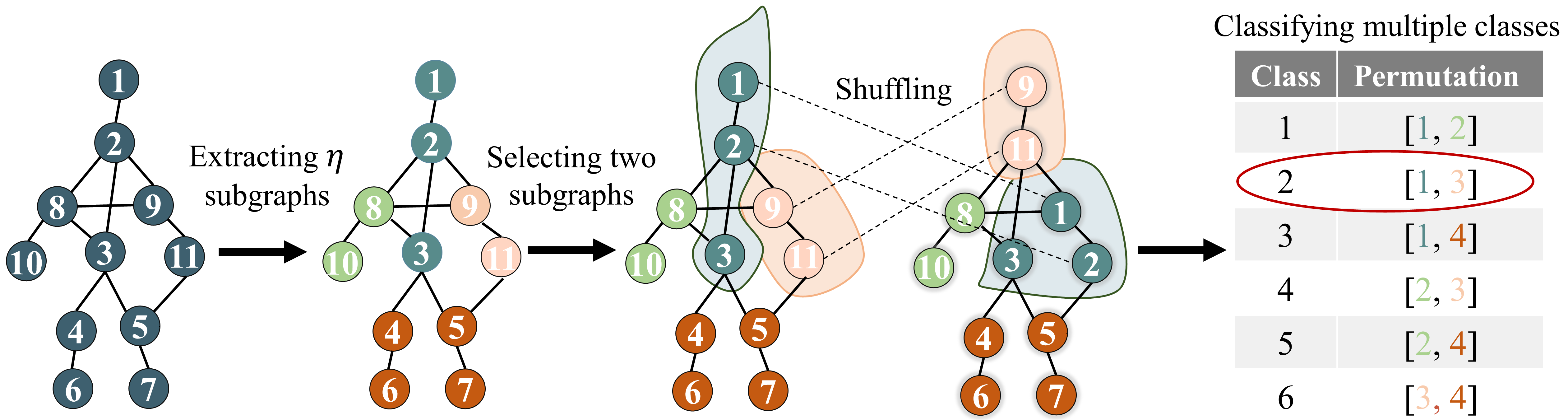}

   \caption{The overview of the graph-level Jigsaw puzzle-solving approach. Nodes with the same color formulate a subgraph. Note that although each node is required to have $\delta$ connections, for better visualization, this property is not strictly maintained in the figure.}
   \label{fig:puzzle_maker}
\end{figure}

\br{In self-supervised learning, the quality of pseudo-labeled data plays a pivotal role in the effectiveness of the learning process. It is essential to curate the pseudo-labeled data that is neither ambiguous nor too easy for the model to solve \cite{noroozi2016unsupervised,wang2022video}. As mentioned earlier in the Introduction, subtle differences between normal and abnormal actions can oftentimes be localized to specific regions of the body rather than affecting the entire body. \s{\b{To incorporate the concept of region-specific discrepancies into the model,}}Therefore, we introduce a novel graph-based Jigsaw puzzle-solving approach as a self-supervised learning method, shown in \cref{fig:puzzle_maker}. Note that while the Jigsaw puzzle-solving task has been widely used in the context of the image domain \cite{misra2020self,taleb2021multimodal,taleb20203d,bucci2021self}, our adaptation marks a pioneering step into the realm of graphs.}

\br{To represent the notation of body regions, we initially partition the graph $\mathcal{A}$, learned in \cref{sec:graph_attention}, into subgraphs. Here, the entire graph corresponds to the entire body, while each subgraph corresponds to an individual body region. However, subgraph identification itself is very challenging as we aim to extract subgraphs that are as distinct as possible from each other while maintaining a close relationship among nodes within the same subgraph. To this end, we employ a subgraph extraction algorithm, namely the Girvan-Newman algorithm \cite{girvan2002community} to extract $\eta$ subgraphs from the adjacency matrix $\mathcal{A}$. Note that these subgraphs may not have the same size, i.e., they could have different numbers of nodes. Then, we select two of these $\eta$ subgraphs randomly and swap their nodes and connections, creating a perturbed version of the adjacency matrix called $\mathcal{A'}$. However, unlike traditional image-level puzzle-solving approaches, shuffling graph-level puzzles is very challenging and requires careful consideration. }

\br{First, while in the image-level Jigsaw puzzling approach, the definition of each puzzle remains constant across all images, our task involves learning the adjacency matrix that evolves over time (see \cref{eq:adj_matrix}).
As a result, the subgraphs that we identify may change dynamically over time. Additionally, the number of nodes within each subgraph may vary, making it challenging to shuffle the positions of subgraphs. 
Unlike image-level puzzles where shuffling merely permutes the locations of puzzle pieces, in graph-level puzzles, shuffling two subgraphs of different sizes, alters the intra-connections within the bigger subgraph. For example, as illustrated in \cref{fig:puzzle_maker}, the larger teal subgraph and the smaller pink subgraph are selected to be shuffled. Upon shuffling, while the pink subgraph maintains its internal connections, the connections within the teal subgraph undergo significant changes, i.e., nodes \#1 and \#2 are no longer connected to node \#3 after shuffling. This is due to the fact that nodes \#9 and \#11 were not connected to node \#3 before shuffling; hence after shuffling, nodes \#1 and \#2 would not be connected to node \#3 as well. Note that after shuffling, nodes \#1, \#2, and \#3 are still considered part of the same subgraph, which deviates the original definition of subgraphs, as they lack a strong internal connection within the subgraph. \brown{This stands as an example of needing an effective subgraph shuffling strategy.}}

\br{Therefore, we present an effective shuffling mechanism. In this process, upon the random selection of two subgraphs, we initiate the shuffling procedure by swapping the most densely connected node in each subgraph (e.g., swap node \#2 in the teal subgraph with node \#11 in the pink subgraph). Here, we determine a node's density by counting the number of intra-connections that it has with other nodes in the same subgraph (e.g., node \#2 has two connections). As per our graph definition in \cref{sec:graph_attention}, all nodes possess $\delta$ connections; we exclusively consider intra-subgraph connections when computing node density. This avoids the trivial solution where all nodes have the same density, which is equal to $\delta$.} The process continues by swapping the next most dense node in each subgraph with each other until the last node is shuffled. The rationale behind prioritizing the shuffling of the most dense nodes \t{is to increase the possibility that they stay connected after shuffling as well. This is due to the understanding that these dense nodes play a crucial role in preserving the overall structure of subgraphs. \s{of preserving the connectivity structure of these nodes, \r{knowing that the top dense nodes in both subgraphs are more probably connected}}}

After shuffling, the model is tasked with a multi-class classification problem where each possible permutation is considered as a class. Our objective is to identify which of the two subgraphs have been shuffled. To this end, we use $\mathcal{H}$, obtained from the first module, project it into the subgraph head, which is a fully-connected layer in our implementation, with the output size of the number of classes, and obtain $\hat{p}$.  Hence, the loss of the Graph-level Jigsaw Puzzle Maker module is the cross-entropy (CE) described below:

\begin{equation} \label{eq:loss_puzzle}
    \mathcal{L}_{\text{puzzle}} = \sum_{z=1}^{\binom{\eta}{2}} \text{CE}(p_z, \hat{p}_z),
\end{equation}
where $p_z$ and $\hat{p}_z$ are the ground truth of the selected class and the predicted probability of each class, respectively.

\br{This task forces the model to develop a better understanding of how each body region contributes to the overall normal behavior. Indeed, each class of shuffling can be interpreted as a form of structural augmentation applied to normal data. While these augmentations may be viewed as variations in the normal structure, which our proposed Graph Attention-based Forecasting module can potentially learn, we aim to classify these structural augmentations through $\mathcal{L}_\text{puzzle}$ as well, to repel the latent space of each augmentation from others. Thus, by jointly optimizing both $\mathcal{L}_\text{graph}$ and $\mathcal{L}_\text{puzzle}$, our model not only learns from the additional structural augmentations applied to normal data but also learns the latent space for each augmentation that is compact and well-separated from other latent spaces of augmentations. 
This allows the reserve of the inter-augmentation spaces for potential abnormal samples that could be observed in the test phase. In essence, during testing, if a sample does not align well with any of the learned augmentation regions, it may be indicative of an anomaly. }

\subsection{Graph-based Conditional Diffusion Model} \label{sec:diffusion}
As mentioned in the Introduction, there are infinite variations of normal and abnormal behavior in skeleton-based video data. Put differently, there are infinitely many normal and anomalous ways of executing an action, each characterized by subtle variations in movement, timing, and context. Conventional approaches \cite{luo2021normal, markovitz2020graph, morais2019learning} that aim to learn a single reconstruction of the input or a singular representation of normal actions would not be able to capture the wide spectrum of human motions. To address this issue, in this module, we propose to utilize diffusion-based techniques \cite{ho2020denoising,wang2022video}, which can generate diverse samples from noise. This provides a more comprehensive exploration of the data space, making anomaly detection algorithms more robust to variations \br{in the normal data during the training phase, as well as both normal and abnormal actions in the inference phase}. The background on diffusion models is provided in Section C of the Supplementary Material.

Our objective is to generate a diverse set of reconstructions of the noisy future frames conditioned on $\mathcal{H}$ learned from the past frames (see \cref{sec:graph_attention}). Essentially, a diffusion model incorporates two Markov chains: a forward process and a reverse process. Given an original clean motion sequence, which is future frames $\mathbf{x}^{+}$, the forward diffusion process incrementally corrupts the coordinates of the joints over a predefined number of steps $T$, making them indistinguishable from a pose with spatial coordinates of joints sampled at random. We sample the noise as $\epsilon^{l+1:L} \in \mathbb{R}^{(L-l)\times K}$ from a normal distribution $\mathcal{N}(0,\mathbf{I})$, where $\mathbf{I}$ is the identity matrix. For simplicity, we represent $\epsilon^{l+1:L}$ as $\epsilon$. The magnitude of the added noise depends on a variance scheduler $\beta_t \in (0,1)$, which controls the quantity of noise added at the $t$-th diffusion step.

The reverse diffusion process, which is the focus of our model's learning, is defined by its capability to reconstruct the original future frames from their noised versions. To estimate the noise, we employ a U-Net-shaped stack of STS-GCN \cite{sofianos2021space} layers, which is capable of capturing the spatio-temporal dependencies of human joints. We train the network conditioned on the embedded multi-layer perception (MLP) of the diffusion step, i.e., $f_\psi(t)$, and $\mathcal{H}$. Inspired by \cite{rombach2022high, flaborea2023multimodal}, the loss is derived as: 

\begin{equation}
\brown{\mathcal{L}_\text{noise} = \mathbb{E}_{\mathbf{x}^{+}, \epsilon, t}\Big[||\epsilon - \epsilon_{\psi}(\mathbf{x}_t^{+}, f_\psi(t), \mathcal{H})||\Big],}
\end{equation}
where $\mathbb{E}$ denotes the expectation, and the term $\epsilon_{\psi}(\mathbf{x}_t^{+}, f_\psi(t), \mathcal{H})$ denotes the U-Net model's estimation of the added noise at the $t$-th time step.

Following \cite{girshick2015fast}, to have a more stable optimization process, we obtain the loss of the Graph-based Conditional Diffusion Model by smoothing $\mathcal{L}_{\text{noise}}$ as follows:

\begin{equation}
\label{eq:smooth}
\mathcal{L}_{\text{diffusion}} = 
\begin{cases} 
0.5 \cdot (\mathcal{L}_\text{noise})^2 & \text{if } |\mathcal{L}_\text{noise}| < 1 \\
|\mathcal{L}_\text{noise}| - 0.5 & \text{otherwise}
\end{cases}
\end{equation}

Thus, the total loss in the training phase is:

\begin{equation}
    \mathcal{L} = \lambda_1 (\mathcal{L}_\text{graph}+\lambda_2\mathcal{L}_\text{puzzle}) +  \mathcal{L}_{\text{diffusion}},
\end{equation}
where $\lambda_1$ and $\lambda_2$ are hyperparameters defined to weigh the importance of each module.

During the inference phase, GiCiSAD predicts future motion frames based on the observed past frames. 
% Subsequently, it aggregates these predictions statistically to detect anomalies. 
%
The process starts from a completely noised state $\mathbf{u}^{+}_T$, randomly drawn from $\mathcal{N}(0,\mathbf{I})$ and continues by iteratively computing $\mathbf{u}^{+}_{t-1}$ from $\mathbf{u}^{+}_{t}$ for $t=T, T-1, \ldots, 1$. The reverse process is then presented as:

{\small
\begin{equation}
\mathbf{u}^{+}_{t-1} = \frac{1}{\sqrt{1-\beta_t}}\left(\mathbf{u}^{+}_t - \frac{\beta_t}{\sqrt{1-\Bar{\alpha_t}}} \epsilon_\psi(\mathbf{u}^{+}_t, f_{\psi}(t), \mathcal{H})\right) + \xi \sqrt{\beta_t},
\end{equation}
}
where $\Bar{\alpha_t} = \prod_{\gamma=T}^{t} (1 - \beta_\gamma$) and $\xi \sim \mathcal{N}(0,\mathbf{I})$.

\br{Motivated by infinite variations of performing both normal and abnormal actions,} we generate $M$ diverse sets of future frames, i.e., $\mathbf{u}^{+}_{(1)}, \mathbf{u}^{+}_{(2)}, \ldots, \mathbf{u}^{+}_{(M)}$. For each generation $m$, where $m \in M$, we compute the reconstruction error, i.e., $\mathcal{S}_m = \mathcal{L}_{\text{diffusion}} (|\mathbf{x}^{+} - \mathbf{u}^{+}_{(m)}|)$ and consider it as the anomaly score. To aggregate anomaly scores of all $M$ generations, we consider three strategies: the mean, the median, and the minimum distance selector. In the mean and median approaches, we derive either the mean or the median of all $M$ scores and allocate this value to the respective frame to evaluate its anomaly level. Regarding the minimum distance selector approach, the lowest anomaly score among all scores is assigned to the respective frame. Note that in the case of more than one actor performing in the scene, the average anomaly score over all actors is assigned to those frames. Our experiments demonstrate that the minimum distance strategy consistently yields the best results, which is consistent with the findings shown in \cite{flaborea2023multimodal}. Details of the experimental results for other aggregation strategies are shown in Section D of the Supplementary Material.

\section{Experiments}
\subsection{Experimental Settings}
\s{In this section, we introduce the datasets, the implementation details, the baselines, and the evaluation metric. }

\textbf{Datasets.} We use four widely used datasets in SVAD, namely Human-Related (HR) versions of the ShanghaiTech Campus (HR-STC) \cite{luo2017revisit}, HR-Avenue \cite{lu2013abnormal}, UBnormal \cite{acsintoae2022ubnormal} and HR-UBnormal \cite{flaborea2023contracting}. HR-STC consists of 13 scenes recorded by different cameras. It contains a total of 303 training videos and 101 test videos with 130 anomalous events. HR-Avenue is comprised of 16 training and 21 test videos with a total of 47 anomalous events. UBnormal comprises 29 scenes generated from 2D natural images using Cinema4D software with 186 normal training and 211 test videos that include 22 categories
of anomalies. We also evaluate our model on a subset of UBnormal, called HR-UBnormal, focusing on only human-related anomalies. This excludes 2.32$\%$ frames, which are non-human-related anomalies, from the test set.

\textbf{Implementation Details.} To ensure a fair comparison with our latest competitors \cite{flaborea2023multimodal, stergiou2024holistic}, we employ a window size of six frames (i.e., $L=6$) for all the experiments, where the first three frames (i.e., $l=3$) are used as the past frames inputted into the Graph Attention-based Forecasting module, while the subsequent three frames are used as the future frames and are fed into the Graph-based Conditional Diffusion Model. For the Graph Attention-based Forecasting module, hyperparameters $D$ and $\delta$ are set to 16 and 5, respectively. The hidden layer in the output prediction of the Graph Attention-based Forecasting module has $128$ neurons. \brown{For the Graph-based Conditional Diffusion Model, we set $\beta_1 = 1e^{-4}$ and $\beta_T = 0.01$, $T = 10$ and apply the cosine variance scheduler from \cite{nichol2021improved}. }  Inspired by \cite{flaborea2023multimodal}, our U-Net reduces the number of joints from $17$ to $10$ and changes the channels from 2 to $(32, 32, 64, 64, 128, 64)$. The diffusion time steps are encoded using the encoding mechanism described in \cite{vaswani2017attention}. We assign values of $\lambda_1 = 0.01$ and $\lambda_2 = 1$. Adam optimizer \cite{kingma2014adam} is utilized with a learning rate set at $10^{-4}$. The batch size of the HR-Avenue dataset is $1024$ and $2024$ for the other datasets. Note that for a fair comparison, the above hyperparameters are fixed in all our experiments and are the same across all datasets.

\textbf{Baselines.} We compare GiCiSAD against the most recent SOTA methods from the literature, including GEPC \cite{markovitz2020graph}, PoseCVAE \cite{jain2021posecvae}, STGCN-LSTM \cite{li2022human}, \s{SSMTL++ \cite{barbalau2023ssmtl++},}COSKAD \cite{flaborea2023contracting}, MocoDAD \cite{flaborea2023multimodal}, and TrajREC \cite{stergiou2024holistic}. Details of the baselines are given in Section E of the Supplementary Material.

\setlength{\tabcolsep}{2.5pt}
\begin{table*}[tb]
\centering
\scalebox{0.7}{
\resizebox{\textwidth}{!}{%
\begin{tabular}{@{}lcccccc@{}}
\toprule
\textbf{Method} & \textbf{Venue} & \textbf{HR-STC} & \textbf{HR-Avenue} & \textbf{HR-UBnormal} & \textbf{UBnormal} \\ \midrule 
% Conv-AE \cite{hasan2016learning} & CVPR 2016 & 69.8 & 84.8 & - & - \\
% Frame-Pred \cite{liu2018future} & CVPR 2018 & 72.7 & 86.2 & - & - \\
% MPED-RNN \cite{morais2019learning}  & CVPR 2019 & 75.4 & 86.3 & 61.2 & 60.6 \\
GEPC \cite{markovitz2020graph} & CVPR 2020 & 74.8 & 58.1 & 55.2 & 53.4 \\
% Multi-Pred\s{Multi-timescale Prediction} \cite{rodrigues2020multi} & WACV 2020 & 77.0 & 88.3 & - & - \\
% Normal Graph \cite{luo2021normal} & Neurocomputing 2021 & 76.5 & 87.3 & - & - \\
PoseCVAE \cite{jain2021posecvae} & ICPR 2021 & 75.7 & 87.8 & - & - \\
% BiPOCO \cite{kanu2022bipoco} & arXiv 2022 & 74.9 & 87.0 & 52.3 & 50.7 \\
STGCN-LSTM  \cite{li2022human} & Neurocomputing 2022 & 77.2 & 86.3 & - & - \\
% SSMTL++ \cite{barbalau2023ssmtl++} & CVIU 2023 & - & - & - & 62.1 \\
COSKAD \cite{flaborea2023contracting} & arXiv 2023 & 77.1 & 87.8 & 65.5 & 65.0 \\
MoCoDAD \cite{flaborea2023multimodal} & ICCV 2023 & 77.6 & 89.0 & 68.4$^{\dagger}$ & 68.3$^{\dagger}$ \\
TrajREC \cite{stergiou2024holistic} & WACV 2024 & 77.9$^{\dagger}$ & 89.4$^{\dagger}$ & 68.2 & 68.0 \\ \hdashline
\textbf{Ours} & - & \textbf{78.0} & \textbf{89.6} & \textbf{68.8} & \textbf{68.6} \\\bottomrule
\end{tabular}%
}}
\caption{Comparison between existing methods and GiCiSAD. The best
and second-best AUROC scores are denoted in bold and $^{\dagger}$.}
\label{tab:main_results}
\end{table*}

\textbf{Evaluation Metric.} Following the common practice in the VAD field \cite{acsintoae2022ubnormal, barbalau2023ssmtl++, flaborea2023multimodal, li2022human, kanu2022bipoco, jain2021posecvae}, we report the Receiver Operating Characteristic Area Under the Curve (AUROC) to assess the performance of our proposed GiCiSAD method.

\subsection{Comparison with State-of-The-Art}

The performances of GiCiSAD and existing methods are summarized in \cref{tab:main_results}. GiCiSAD demonstrates superior performance with the AUROC scores of 78.0, 89.6, 68.8, and 68.6 on HR-STC, HR-Avenue, HR-UBnormal, and UBnormal datasets, respectively. GiCiSAD outperforms all existing methods, including the most recent competitors, MoCoDAD and TrajREC. This can be attributed to GiCiSAD's comprehensive approach in addressing critical challenges prevalent in SVAD. As outlined in the Introduction, skeleton-based video data poses several critical challenges, including spatio-temporal dependencies, region-specific discrepancies, and infinite variations. While previous methods have attempted to tackle individual aspects of these challenges, none have provided a comprehensive solution. For instance, prior methods such as \cite{markovitz2020graph,luo2021normal,li2022human,flaborea2023contracting} have leveraged graph-based models to capture spatio-temporal dependencies, yet they failed to take into account other challenges, such as region-specific discrepancies, and infinite variations. Similarly, while \cite{flaborea2023multimodal} effectively tackled the issue of infinite variations through a conditional diffusion-based model, it neglected others. \s{Moreover, the exploration of SSL in SVAD has been limited, with only one study \cite{barbalau2023ssmtl++} identified in the literature}Moreover, none of the existing methods have explicitly addressed the challenge of region-specific discrepancies. In contrast, GiCiSAD stands out as it systematically addresses all these challenges through three novel proposed modules, each specifically designed for a specific issue, leading to improved detection performance.

\subsection{Parameter Efficiency}

\cref{tab:params} presents a comparative analysis of GiCiSAD and \t{the most recent unsupervised competitors, i.e., MoCoDAD and TrajREC, as well as two recently developed supervised methods, i.e., AED-SSMTL and TimeSformer. Note that supervised methods have access to both normal and abnormal data during the training phase, while unsupervised methods have only access to the normal data in their training phase. The results show that our method outperforms the unsupervised methods from the anomaly detection accuracy point of view. Also, our model shows a significant reduction of the number of parameters — up to 40\% less than the most parameter-efficient unsupervised method, MoCoDAD. Moreover, compared with the supervised methods, we achieve comparable performance in terms of AUROC, with only a fraction of the parameters of their models.}

% regarding anomaly detection performance and the number of training parameters. It is shown that in terms of anomaly detection performance, our method outperforms all existing unsupervised methods and our method is competitive with the existing supervised approaches \cite{acsintoae2022ubnormal}, which rely on the true labeled data during training. This demonstrates the efficacy of our self-supervised method that achieves superior performance without the need for the true labeled data. Additionally, GiCiSAD has a significantly lower number of parameters compared to existing methods. For example, our model consists of only a fraction of the parameters found in supervised methods. Also, compared to the most recent and parameter-efficient unsupervised method, i.e., MoCoDAD, our model reduces the number of parameters up to 40\%, underscoring its efficiency and scalability. This dual advantage positions GiCiSAD as a leading solution in the realm of SVAD.

\begin{table}[tb]
\centering
\scalebox{0.9}{
\begin{tabular}{lcc}
\toprule
\textbf{Method} \hspace{1em} & \textbf{Params} \hspace{1em} & \textbf{AUROC} \\
\midrule 
AED-SSMTL$^{\diamondsuit}$ \cite{georgescu2021background}   & $>$80M & 61.3 \\
TimeSformer$^{\diamondsuit}$ \cite{bertasius2021space} & 121M & \textbf{68.6} \\[0.5ex]
\hdashline
\\[-2.5ex]
TrajREC & 4.9M & 68.0 \\
MoCoDAD & 142K$^{\dagger}$ & 68.3$^{\dagger}$ \\
\textbf{GiCiSAD} & \textbf{82.6K} & \textbf{68.6} \\
\bottomrule
\end{tabular}}
\caption{Comparison between GiCiSAD and existing methods in terms of AUROC on the UBnormal dataset and the number of training parameters (Params). $^{\diamondsuit}$ denotes the supervised methods. The best and second-best results are denoted in bold and $^{\dagger}$.}
\label{tab:params}
\end{table}

\subsection{Ablation Study}
We assess the effectiveness of GiCiSAD through ablation studies on HR-Avenue and HR-STC datasets considering four key aspects of GiCiSAD: (1) effectiveness of individual components, (2) effectiveness of conditioning mechanism, (3) types of graph-based Jigsaw puzzles, and (4) the number of subgraphs. For simplicity, we refer to Graph Attention-based Forecasting, Graph-based Jigsaw Puzzle Maker, and Graph-based Conditional Diffusion Model as \texttt{Graph}, \texttt{Puzzle}, and \texttt{Diffusion}, respectively.

\textbf{Effectiveness of Individual Components}. As described in our inference phase, GiCiSAD utilizes anomaly scores derived from the reverse process of \texttt{Diffusion}; hence, its inclusion is essential for anomaly detection. Therefore, in this experiment, we conduct ablation studies specifically focusing on the exclusion of either \texttt{Graph} or \texttt{Puzzle}. The results are shown in \cref{tab:individual}. The sign ``+'' denotes the inclusion of a component. Results show that each component in our model plays a crucial role in improving its performance. For instance, in the scenario where we only include \texttt{Graph} and \texttt{Diffusion}, our model achieves a 77.4\% AUROC on HR-STC, which proves our hypothesis that despite the capability of capturing the overall spatio-temporal dependencies inherent in skeleton-based data, \texttt{Graph} itself lacks a deeper understanding of the impact of each region of the human body. Conversely, if we include \texttt{Puzzle} and \texttt{Diffusion}, our model can capture the region-specific discrepancies, yet it is not able to understand the overall nature of human actions and thus, adding another objective that stands for capturing the overall normal structure of the body is needed. Finally, when we include all components together, our model achieves the best result with an AUROC of 78.0\% as it addresses all the challenges comprehensively.

\begin{table}[tb]
\centering
\scalebox{0.9}{
\begin{tabular}{lcc}
\toprule
                    \textbf{GiCiSAD}    & \textbf{HR-Avenue} \hspace{1em} & \textbf{HR-STC} \\ \midrule
\texttt{Graph} + \texttt{Diffusion}     & 88.2        & 77.4     \\
\texttt{Puzzle} + \texttt{Diffusion}     & 87.9        & 77.2     \\
\texttt{Graph} + \texttt{Puzzle} + \texttt{Diffusion}           & \textbf{89.6}        & \textbf{78.0} \\ \bottomrule
\end{tabular}}
\caption{The performance of individual components and their combination in GiCiSAD.}
\label{tab:individual}
\end{table}

\textbf{Effectiveness of Conditioning Mechanism}.
In this study, we experiment with three different conditioning strategies to be used in \texttt{Diffusion}, including our proposed \textit{Graph}-based, introduced in \cref{sec:graph_attention}, \textit{Encoder}-based and \textit{AutoEncoder}-based conditioning mechanisms. \t{The architecture of the two latter ones is borrowed from our main competitor, MoCoDAD \cite{flaborea2023multimodal}}. The objective is to evaluate which of these approaches yields the most effective latent representation from past frames to guide \texttt{Diffusion}. In the \textit{Encoder}-based approach, the encoder architecture constructs the conditioning latent space for \texttt{Diffusion} without introducing an additional loss to the network. In the \textit{AutoEncoder}-based approach, the reconstruction loss of the autoencoder is added to the loss of \texttt{Diffusion}. More details of their architectures are provided in Section F of the Supplementary Material. For a fair comparison, since in \textit{Encoder}-based and \textit{AutoEncoder}-based approaches, no graphs are constructed to build the Jigsaw puzzles upon, in our proposed \textit{Graph}-based approach, we exclude \texttt{Puzzle}, and keep only \texttt{Graph} and \texttt{Diffusion}. \cref{tab:conditioning} indicates that
in general, the \textit{Autoencoder}-based approach outperforms \textit{Encoder}-based approach. This discrepancy can be attributed to the supervision of the reconstruction loss in the \textit{Autoencoder}-based method, which aids in obtaining a better representation of past frames. In contrast, our proposed \textit{Graph}-based method achieves significantly higher AUROC scores. This is due to \texttt{Graph}'s ability to effectively capture spatio-temporal dependencies in the data. Parameter-wise, in our model, each joint is only connected to a few other joins, while the \textit{Encoder}-based and \textit{Autoencoder}-based methods are fully connected neural networks, where all joints contribute to the output. This results in a much lower number of parameters in our proposed method compared to those other methods.

% considers a subset of nodes, where each node only has $\delta$ connections, parameter wise

% \r{Specifically, each joint exhibits unique behavior and is dependent on only a few other joints.} \s{Thus, dynamically modeling this behavior is necessary.}  On the other hand, the \textit{Encoder}-based and \textit{Autoencoder}-based methods are fully-connected neural networks, where all joints contribute to the output. This results in a much lower number of parameters in our proposed method compared to these two methods.

% we achieve an AUROC of 86.8\% in HR-Avenue with the \textit{Autoencoder}-based approach, whereas the \textit{Encoder}-based approach yields only 83.9\%. This discrepancy can be attributed to the supervision of the reconstruction loss in the \textit{Autoencoder}-based method, which aids in obtaining a better representation of past clean frames. In contrast, our proposed \textit{Graph}-based method achieves a significantly higher AUROC score of 88.2\%. This is due to \texttt{Graph}'s ability to effectively capture spatio-temporal dependencies in the data. Specifically, each joint exhibits unique behavior and is dependent on only a few other joints. Thus, dynamically modeling this behavior is necessary.  On the other hand, the \textit{Encoder}-based and \textit{Autoencoder}-based methods are based on fully-connected neural networks, where all joints contribute to the output. This results in a much lower number of parameters in our proposed method compared to these two methods.

\begin{table}[tb]
\centering
\scalebox{0.9}{
\begin{tabular}{lccc}
\toprule
                    \textbf{Conditioning Mechanism} \hspace{0.5em}    & \textbf{HR-Avenue} \hspace{0.5em} & \textbf{HR-STC} \hspace{0.5em} & \textbf{Params}  \\ \midrule
\textit{Encoder}-based     & 83.9        & 74  &  111.1K \\
\textit{AutoEncoder}-based          & 86.8        & 76.6 & 142.3K\\ 
\textit{Graph}-based     & \textbf{88.2}        & \textbf{77.4}  &  \textbf{82.6K} \\ \bottomrule
\end{tabular}}
\caption{The performance of different conditioning mechanisms for \texttt{Diffusion}.}
\label{tab:conditioning}
\end{table}

\textbf{Types of Graph-based Jigsaw Puzzles}. In this experiment, we explore the efficacy of various graph-based Jigsaw puzzling strategies. Specifically, we compare our proposed Jigsaw puzzling method described in \cref{sec:puzzel_maker}, so-called \textit{Inter-community}, which selects two subgraphs and interchanges them, with a new strategy, called \textit{Intra}-community. This new technique chooses a single subgraph and randomly rearranges the nodes inside, where the objective is to determine which of these subgraphs has been changed.  A visualization of this technique is provided in Section G of the Supplementary Material. As shown in  \cref{tab:puzzle},  \textit{Inter}-community yields superior results compared to \textit{Intra}-community. This could be attributed to our method of subgraph identification, which aims to select subgraphs that are as distinct from each other as possible while preserving the tight connections among nodes within the same subgraph. Consequently, shuffling within the same subgraph is less effective because the nodes are highly interdependent, and detecting the shuffled subgraph \t{is a simpler pretext task compared to the \textit{Inter}-community shuffling, hence, less guidance is provided through the self-supervision process.} \s{In contrast, shuffling between two distinct subgraphs, each representing different region of the human body, enhances the model's understanding of how these sub-components contribute to the body's overall functionality.}

% \r{Combining these strategies, denoted as \textit{Both}, yields better results than \textit{Intra}-community alone, but the ambiguity introduced by \textit{Intra}-community prevents the model from achieving the same performance as \textit{Inter}-community alone. Although we have gained some insights, further exploration on this research area is necessary.}

\begin{table}[tb]
\centering
\scalebox{0.9}{
\begin{tabular}{lcc}
\toprule
                        & \textbf{HR-Avenue} \hspace{1em} & \textbf{HR-STC} \\ \midrule
\textit{Inter}-Community      & \textbf{89.6}        & \textbf{78.0}    \\
\textit{Intra}-Community           & 88.5        & 77.5     \\
% \textit{Both}  & 89.2        & 77.7     \\ \bottomrule
\bottomrule
\end{tabular}}
\caption{Different graph-based Jigsaw puzzling strategies.}
\label{tab:puzzle}
\end{table}

\begin{table}[tb]
\centering
\scalebox{0.9}{
\begin{tabular}{lcc}
\toprule
                     \textbf{$\eta$} \hspace{1em}  & \textbf{HR-Avenue} \hspace{1em} & \textbf{HR-STC} \\ \midrule
2    &    88.6     &   77.5  \\  
3     &    88.9    &   77.7   \\                   
4     & \textbf{89.6}      & 77.9     \\
5     & 89.5        & \textbf{78.0}     \\
6          & 89.5        & 77.9 \\ 
7     &      89.3   &    77.8  \\  \bottomrule
\end{tabular}}
\caption{Variability on the number of subgraphs.}
\label{tab:eta}
\end{table}

\textbf{Number of Subgraphs.} \brown{This experiment varies the number of subgraphs, denoted as $\eta$, to be extracted from the graph. Results are shown in \cref{tab:eta}. We range the value of $\eta$ from 2, where the constructed subgraphs are relatively small, up to 7, where the constructed subgraphs become too large. Our results suggest that slight variations do not adversely affect the model's overall performance as long as the size of subgraphs remains reasonable to adequately represent body regions, with $\eta$ values between 4 and 6.}

\section{Conclusion}

In this study, we introduce GiCiSAD, a novel and lightweight framework designed to effectively tackle three critical challenges encountered in SVAD datasets. We emphasize the importance of a challenging pretext task to learn a discriminative representation of region-specific discrepancies between normal and abnormal motions, necessitating a dynamic graph-based modeling of human body motions. Additionally, we introduce a novel conditional diffusion model to generate a wide spectrum of future human motions guided by the encoded representations of past frames through learnable graphs. Experimental results validate the efficacy of our approach, showcasing SOTA performance on four popular benchmarks while employing significantly fewer training parameters compared to the SOTA.

%%%%%%%%%%%%

%%%%%%%%%%%%%%%%%%%%%%%%%%%%%%%%%%%

%%%%%%%%%%%%%%%%%%%%%%%%%%%%%%%%%%%%%

%%%%%%%%% REFERENCES
{\small
\bibliographystyle{ieee_fullname}
\bibliography{egbib}
}

\clearpage  % Start a new page for the appendix

\appendix  % Switch to appendix mode, changing section numbering

\begin{center}
    \Large \textbf{APPENDIX}  % Use a larger font size and bold text for the title
\end{center}

\vspace{1em} 

This appendix provides supplementary details for the WACV 2025 paper titled \textit{"Graph-Jigsaw Conditioned Diffusion Model for Skeleton-based Video Anomaly Detection"}.

\begin{itemize}
    \item \cref{sec:Related_Work1} provides a comprehensive reviews of related works.
    \item \cref{sec:GaLaSAD_Algorithms} presents the pseudocode for our proposed GiCiSAD method.
    \item  \cref{sec:Background_Diffusion_Models} provides the background of diffusion models.
    \item \cref{sec:aggregations} presents the results for different statistical aggregations of anomaly scores in the inference phase.
    \item  \cref{sec:Baselines} provides information of the baselines.
    \item \cref{sec:Conditioning_Mechanism} provides information of different conditioning strategies.
    \item \cref{sec:Jigsaw_Puzzles_type} provide a visualization of the \textit{Intra}-community shuffling approach.
    \item \cref{sec:visualization} illustrates an example of the \textit{Inter}-community shuffling approach on the real-world constructed graph.
\end{itemize}

\section{Related Work}
\label{sec:Related_Work1}

\subsection{Skeleton-based Video Anomaly Detection}

Skeleton-based video anomaly detection (SVAD) has gained significant attention in recent years due to its potential applications in various domains such as video surveillance, healthcare, and human-computer interaction. Many studies have leveraged the powerful representation capabilities of deep learning to automatically learn features from skeleton-based video data, hence, to improve the anomaly detection performance. Existing deep learning studies can be categorized into three main approaches \cite{wang2022video,mishra2024skeletal}: reconstruction-based, prediction-based and hybrid approaches. In the reconstruction-based approach \cite{gong2019memorizing,nguyen2019anomaly}, an autoencoder or its variant model is trained on only normal human activities. During training, the model learns to reconstruct the samples representing normal activities, hence it is expected to yield low reconstruction error for normal data, while achieving high reconstruction error for abnormal data in the test phase. Regarding the prediction-based approach \cite{liu2018future,yu2020cloze,feng2021convolutional}, a model is trained to learn the normal human behaviors by predicting the skeletons at the next time steps using information at past time steps. During the test phase, the test samples
with high prediction errors are flagged as anomalies. Lastly, the combination of reconstruction-based and prediction-based approaches, which is called as the hybrid approach, has been also widely explored \cite{zhao2017spatio,ye2019anopcn,morais2019learning}. These methods utilize a multi-objective loss function that consists of reconstruction and prediction errors to learn the characteristics
of normal skeletons, aimed at identifying skeletons with large errors as anomalies in the test phase.

However, these three approaches encounter several issues that require more advanced methods to tackle. For example, reconstruction-based methods necessitate the availability of normal data during the training phase, leading to an expectation of higher reconstruction errors for abnormal samples. However, this assumption does not always hold in practice; these methods can also generalize well to anomalies, resulting in false negatives \cite{gong2019memorizing}. In prediction-based approaches, determining the optimal prediction horizon for future (or past) events poses a challenge. Moreover, methods relying on future prediction can be sensitive to noise in past data \cite{tang2020integrating}. Even minor alterations in past data can lead to significant variations in predictions, not all of which necessarily indicate anomalies. The combination-based methods include the limitations of the individual learning approaches. It is also challenging to determine the optimum value of combination coefficients (weights) to balance the importance of individual components in a multi-objective loss function. Importantly, in skeleton-based video data, subtle differences between normal and abnormal actions can oftentimes be localized to specific regions of the body rather than affecting the entire body. However, all existing reconstruction-based, prediction-based and hybrid methods are based on modeling the human body as a whole and ignore the importance of such local variations when detecting anomalies. Note that skeleton-based video data also includes the challenge of inifinite variations of performing normal and abnormal actions. While few studies
has addressed this diversity challenge \cite{flaborea2023multimodal}, by considering the body as whole, they overlook the fact that abnormalities may be localized to only specific regions of the body, potentially leading to misdetection in cases where anomalies occur in isolated regions while the rest of the body remains normal.

\subsection{Self-supervised Learning}

Recently, self-supervised learning (SSL) has been widely employed in the context of video anomaly detection  \cite{wang2022video,huang2022self,hu2023self}. Essentially, SSL leverages large amounts of unlabeled data to learn meaningful representations without requiring explicit annotations for anomaly detection. Not limited to the predefined reconstruction or prediction tasks, SSL methods define various pretext tasks that can adapt to the specific characteristics and complexities of the data, potentially leading to more robust and discriminative representations. Notable approaches include contrastive learning, which learns to maximize agreement between differently augmented views of the same data, as demonstrated by recent works such as SimCLR \cite{chen2020simple} and MoCo \cite{he2020momentum}. Other methods, such as generative adversarial networks (GANs) \cite{chen2021nm,huang2022self} and autoencoders \cite{huang2021self}, have also been explored for self-supervised representation learning from videos. While many studies have demonstrated the capability of SSL, they failed to address the challenge of capturing region-specific features in the field of SVAD. Very few works \cite{wang2022video} have effectively address this challenge by proposing a challenging pretext task, which encourages the model to focus on region-level features in the image domain. However, it remains unanswered how to adapt this approach to the field of SVAD, particularly considering the presence of skeleton data instead of traditional images in this context. This is due to the fact that unlike images, skeleton data exhibits spatial structure, and the temporal dynamics, which both play a crucial role in defining actions and anomalies. Convolutional layers commonly used in image-based SSL may not directly apply to skeleton data. Instead, architectures based on a combination of recurrent neural networks (RNNs) \cite{luo2019video} and graph neural networks (GNNs) \cite{zeng2021hierarchical} can be employed to model the temporal and spatial aspects of skeleton sequences.

\subsection{Graph-based Approaches}

As denoted in the main paper, skeleton data is inherently a time-series data that exhibits sptatio-temporal dependencies. Hence, it can be naturally represented as graphs \cite{markovitz2020graph}, where joints correspond to nodes and the connections between joints form edges. Many studies have exploited the potential of graphs for SVAD tasks. For example, \cite{li2022human} introduces a Spatial-temporal Graph Convolutional Autoencoder with Embedded Long Short-Term Memory Network (STGCAE-LSTM) for SVAD. This architecture comprises a single-encoder-dual-decoder setup capable of simultaneously reconstructing the input and predicting future frames. By leveraging graph convolutional operations, the model captures spatial dependencies among joints. However, its fixed adjacency matrix limits its ability to adapt to evolving relationships between joints over time, potentially hindering its performance in capturing dynamic activities. \cite{luo2021normal} proposes Normal Graph, a spatial-temporal graph convolutional prediction-based network for SVAD. While pioneering in applying graph convolutional networks to SVAD and effectively capturing spatial dependencies, Normal Graph suffers from the same limitation as STGCAE-LSTM in its inability to dynamically learn changing relationships between joints over time, as it fixes the adjacency matrix. Addressing the constraints imposed by fixed adjacency matrices is critical for advancing the state-of-the-art in SVAD. Recent research has explored the capability of dynamically learning graphs overtime in both pure graph and time series domains \cite{deng2021graph,xia2021graph,ho2023graph}. In other words, these models dynamically learn the relationships between nodes over time, offering enhanced capabilities in capturing complex spatio-temporal dependencies and detecting anomalies in dynamic graph or time-series domains. However, to date, there remains a scarcity of works capable of effectively capturing the evolving relationships of joints in real-time skeleton-based video streams.

In response to the limitations observed in existing methodologies within the field, we present GiCiSAD, a comprehensive framework that introduces three novel modules to tackle these challenges effectively. The Graph Attention-based Forecasting module leverages a graph learning strategy to effectively capture the spatio-temporal dependencies. To address the issue of region-specific discrepancies, we propose a novel graph-level SSL with a difficult pretext task, called Graph-level Jigsaw Puzzle Maker, which involves various subgraph augmentations applied to the learnable graph, hence providing supervisory signals to help GiCiSAD capture a slight region-level difference between normal and abnormal behaviors. Lastly, to contend with the infinite variations inherent in anomaly detection tasks, GiCiSAD integrates a cutting-edge diffusion-based model named Graph-level Conditional Diffusion Model. Leveraging the learned graph from previous frames as conditional information, this model generates a diverse array of future samples, thereby enhancing the robustness and adaptability of GiCiSAD.
%%%%%%%%%%%%%%%%%%%%%%%%%%%%%%%

\section{GiCiSAD Pseudocode}
\label{sec:GaLaSAD_Algorithms}

The overall procedure of the training and inference phases of GiCiSAD is described in \cref{alg:train} and  \cref{alg:Inference}, respectively. Note that during the inference phase, $M$ sets of future frames are generated. Subsequently, these generated frames are compared with the actual ground truth, resulting in $M$ anomaly scores. Finally, these scores are consolidated into a single aggregated value. More detail regarding the aggregation mechanism is presented in \cref{sec:aggregations}. In scenarios with more than one actor in the scene, to summarize the anomaly score of all actors, we follow the methodology outlined in \cite{flaborea2023multimodal}. This approach consolidates the contributions of all actors by considering both the average error across all actors and the span of the error range.

\renewcommand{\baselinestretch}{1.1}
\begin{algorithm}[H]
\caption{GiCiSAD Training}
\label{alg:train}
\begin{algorithmic}[1]
\State \textbf{Input:} $X$, diffusion hyperparameters $\{\beta_0, \beta_T, T\}$, $\delta$, $\lambda_1$, $\lambda_2, \eta$.
% \State Compute diffusion hyperparameters $\Bar{\alpha}_t$, $\Bar{\beta}_t$, immediate diffusion steps $t$, and $\epsilon$. 
\State Randomly initialize trainable parameters $\theta$ and $\psi$.

\For {\textit{not converged}}
    \State $[\mathbf{x}^{+},\mathbf{x}^{-}]$ = Batch($X$) \Comment{\textcolor{teal}{Batching}}
    \State Compute $\mathbf{x}^{-}_{avg}$
    \State $\mathcal{A}$ = Graph$_{\theta}$($\mathbf{x}^{-}, \delta$) \Comment{\textcolor{teal}{Adjacency Matrix Calculation}}
    \State $[\mathcal{A}', p]$ = Puzzle($\mathcal{A}, \eta$) \Comment{\textcolor{teal}{Puzzle Making}}
    \State $[\textbf{H}, \hat{\mathbf{x}}^-_{avg}]$ = Attention$_{\theta}$($\mathcal{A}'$) \Comment{\textcolor{teal}{Attention Mechanism}}
    \State $\mathcal{H}$ = FC$_{\theta}$($\mathbf{H}$)
    \State $\hat{p}$ = SubgraphHead$_{\psi}$($\mathcal{H}$)
    \State $[t,\mathbf{x}^+_{\text{corrupted}}, \epsilon]$ = Forward($\mathbf{x}^+, T, \beta_0, \beta_T$)\Comment{\textcolor{teal}{Forward Diffusion}}
    \State $\hat{\epsilon}$ = Reverse$_{\psi}$($\mathbf{x}^+_{\text{corrupted}}, \mathcal{H}, f_{\psi}(t)$) \Comment{\textcolor{teal}{Reverse Diffusion}}
    \State $\mathcal{L}$ = $\lambda_1 \Big(\mathcal{L}_{\text{graph}}(\hat{\mathbf{x}}^-_{avg}, \mathbf{x}^-_{avg}) + \lambda_2 \mathcal{L}_{\text{puzzle}}(\hat{p}, p)\Big) + \mathcal{L}_{\text{diffusion}}(\hat{\epsilon}, \epsilon)$ 
    \State Backpropage $\mathcal{L}$ to update $\theta$ and $\psi$.
\EndFor
% \State \textbf{Return:} $\mathcal{L}$, $\mathbf{u}^+$.

\end{algorithmic}
\end{algorithm}
%\renewcommand{\baselinestretch}{1}
%%%%%%%%%%%%%%%%%%%%%%%%%%%%%%%%%%%%%%%
%\renewcommand{\baselinestretch}{1.1}
\begin{algorithm}
\caption{GiCiSAD Inference}
\label{alg:Inference}
\begin{algorithmic}[H]
\State \textbf{Input:} $\mathbf{x}^{1:L}$, $l$, $\delta$, diffusion hyperparameters $\{\beta_0, \beta_T, T\}$, $\eta$, $M$.
\State $\text{Agg} \leftarrow \emptyset$
\State $\mathbf{x}^-$ = $\mathbf{x}^{1:l}$
\State $\mathbf{x}^+$ = $\mathbf{x}^{l+1:L}$
\State Actors = \{All Actors participating in $\mathbf{x}^{1:L}$\}
\For{a in Actors}\Comment{\textcolor{teal}{Iteration Over Actors}}
    \State $\text{Scores} \leftarrow \emptyset$
    \For{i in range($M$)}\Comment{\textcolor{teal}{Generate $M$ Samples}}
        \State $\mathbf{u}^{+}_i = \mathcal{N}(0,\mathbf{I})$
        \State $\Bar{\alpha} = 1$
        \For{t = T, \ldots, 1}
            \State $\mathcal{A}$ = Graph$_{\theta}$($\mathbf{x}^{-}, \delta$) \Comment{\textcolor{teal}{Adjacency Matrix Calculation}}
            \State $[\mathcal{A}', p]$ = Puzzle($\mathcal{A}, \eta$) \Comment{\textcolor{teal}{Puzzle Making}}
            \State $\mathbf{H}$ = Attention$_{\theta}$($\mathcal{A}'$) \Comment{\textcolor{teal}{Attention Mechanism}}
            \State $\mathcal{H}$ = FC$_{\theta}$($\mathbf{H}$)
            \State $\hat{\epsilon}$ = Reverse$_{\psi}$($\mathbf{u}^{+}_i, \mathcal{H}, f_{\psi}(t)$) \Comment{\textcolor{teal}{Reverse Diffusion}}
    
            % \State -,$\mathbf{u}^+_t$ = GiCiSAD($\mathbf{u}^{+}_i$, $\mathbf{x}^-$, $\delta$, $t$, $\lambda_1$, $\lambda_2, \eta$)\Comment{\textcolor{teal}{Noise Prediction}}
            \State $\xi = \mathcal{N}(0,\mathbf{I})$
            \State $\Bar{\alpha} = \Bar{\alpha} \times (1-\beta_t)$
            \State $\mathbf{u}^{+}_i = \frac{1}{\sqrt{1-\beta_t}}\left(\mathbf{u}^{+}_i - \frac{\beta_t}{\sqrt{1-\Bar{\alpha}}} \hat{\epsilon}\right) + \xi \sqrt{\beta_t}$ \Comment{\textcolor{teal}{Recover The Sequence}}
        \EndFor
        \State $\text{Scores} \leftarrow \text{Scores} \cup \{\mathcal{L}_{\text{diffusion}}(\mathbf{x}^+, \mathbf{u}^+_i)\}$ \Comment{\textcolor{teal}{Save Anomaly Score}}
    \EndFor
    \State $\text{Agg} \leftarrow \text{Agg} \cup \{\text{AGGREGATE}(\text{Scores})\}$ \Comment{\textcolor{teal}{Aggregate $M$ Anomaly Scores}}
\EndFor

\State \textbf{Anomaly Score:} mean(Agg) + $\log\frac{1+\max(\text{Agg})}{1+\min(\text{Agg})}$. \Comment{\textcolor{teal}{Anomaly Score Across All Actors}}

\end{algorithmic}
\end{algorithm}
\renewcommand{\baselinestretch}{1}

\section{Background on Diffusion Models}
\label{sec:Background_Diffusion_Models}

Diffusion models \cite{ho2020denoising,song2020denoising}, a class of generative models, define a two-process paradigm that includes: the forward process that slowly adds Gaussian noise to the data and the reverse process that constructs the desired data from the noise. Mathematically, the forward process incrementally adds Gaussian noise to the initial stage, called $\mathbf{x}_0 \sim q (\mathbf{x}_0)$ over $T$ diffusion steps according to a variance scheduler $\beta_1, \ldots, \beta_T$. The approximate posterior can be represented as:

% \begin{equation}
%     q(\mathbf{x}_{1:T}|\mathbf{x}_0) := \prod_{t=1}^T q(\mathbf{x}_t|\mathbf{x}_{t-1}), \quad q(\mathbf{x}|\mathbf{x}_{t-1}) := \mathcal{N}(\mathbf{x}_t;\sqrt{1-\beta_t}\mathbf{x}_{t-1},\beta_t \mathbf{I})
% \end{equation}
\begin{align}
    q(\mathbf{x}_{1:T}|\mathbf{x}_0) &:= \prod_{t=1}^T q(\mathbf{x}_t|\mathbf{x}_{t-1}), \\
    q(\mathbf{x}_t|\mathbf{x}_{t-1}) &:= \mathcal{N}(\mathbf{x}_t;\sqrt{1-\beta_t}\mathbf{x}_{t-1},\beta_t \mathbf{I})
\end{align}

By setting $\alpha_t := 1-\beta_t$ and $\Bar{\alpha}_t := \prod_{s=1}^t \alpha_s$, the forward process allows to immediately transform $\mathbf{x}_0$ to a noisy $\mathbf{x}_t$ according
to $\beta_t$ in a closed form as:

\begin{equation}
    q(\mathbf{x}_t|\mathbf{x}_0) := \mathcal{N}(\mathbf{x}_t;\sqrt{\Bar{\alpha}_t}\mathbf{x}_0, (1- \Bar{\alpha}_t)\mathbf{I}).
\end{equation}

The reverse process aims to produce the samples that match the data distribution after a finite number of transition steps. Starting with $p(\mathbf{x}_T) := \mathcal{N}(\mathbf{x}_t; 0, \mathbf{I})$, the joint distribution is then given by:

% \begin{equation}
%     p_{\psi}(\mathbf{x}_{0:T}) := p(\mathbf{x}_T) \prod_{t=1}^T p_{\psi}(\mathbf{x}_{t-1}|\mathbf{x}_t), \\
%     \quad p_{\psi}(\mathbf{x}_{t-1}|\mathbf{x}_t) := \mathcal{N}(\mathbf{x}_{t-1}; \mathbf{\mu}_{\psi}(\mathbf{x}_t,t), \mathbf{\sigma}_{\psi}(\mathbf{x}_t,t)).
% \end{equation}
\begin{align}
    p_{\psi}(\mathbf{x}_{0:T}) &:= p(\mathbf{x}_T) \prod_{t=1}^T p_{\psi}(\mathbf{x}_{t-1}|\mathbf{x}_t), \\
    p_{\psi}(\mathbf{x}_{t-1}|\mathbf{x}_t) &:= \mathcal{N}(\mathbf{x}_{t-1}; \mathbf{\mu}_{\psi}(\mathbf{x}_t,t), \mathbf{\sigma}_{\psi}(\mathbf{x}_t,t)).
\end{align}

Note that $\mathbf{\mu}_{\psi}(\mathbf{x}_t,t)$ and $\mathbf{\sigma}_{\psi}(\mathbf{x}_t,t)$ are parameterized as:

% \begin{equation}
%     \mathbf{\mu}_{\psi}(\mathbf{x}_t,t) = \frac{1}{\sqrt{\alpha_t}} \Big(\mathbf{x}_t - \frac{\beta_t}{\sqrt{1-\Bar{\alpha}_t}} \epsilon_{\psi}(\mathbf{x}_t,t)\Big), \quad
%     \mathbf{\sigma}_{\psi}(\mathbf{x}_t,t) = \sqrt{\Bar{\beta}_t},
% \end{equation}
\begin{align}
    \mathbf{\mu}_{\psi}(\mathbf{x}_t,t) &= \frac{1}{\sqrt{\alpha_t}} \Big(\mathbf{x}_t - \frac{\beta_t}{\sqrt{1-\Bar{\alpha}_t}} \epsilon_{\psi}(\mathbf{x}_t,t)\Big), \\
    \mathbf{\sigma}_{\psi}(\mathbf{x}_t,t) &= \sqrt{\Bar{\beta}_t},
\end{align}
where $\Bar{\beta}_t = \frac{1-\Bar{\alpha}_{t-1}}{1-\Bar{\alpha}_t}\beta_t$, and $\Bar{\beta}_1 = \beta_1$. $\epsilon_{\psi}$ is a network approximator (the U-Net-based architecture in our case), which take $\mathbf{x}_t$ and the diffusion step $t$ as the inputs, and aims to predict the noise from $\mathbf{x}_t$.

\section{Different Strategies for Statistical Aggregations}
\label{sec:aggregations}
\begin{figure}[tb]
  \centering
  \includegraphics[width=1\linewidth]{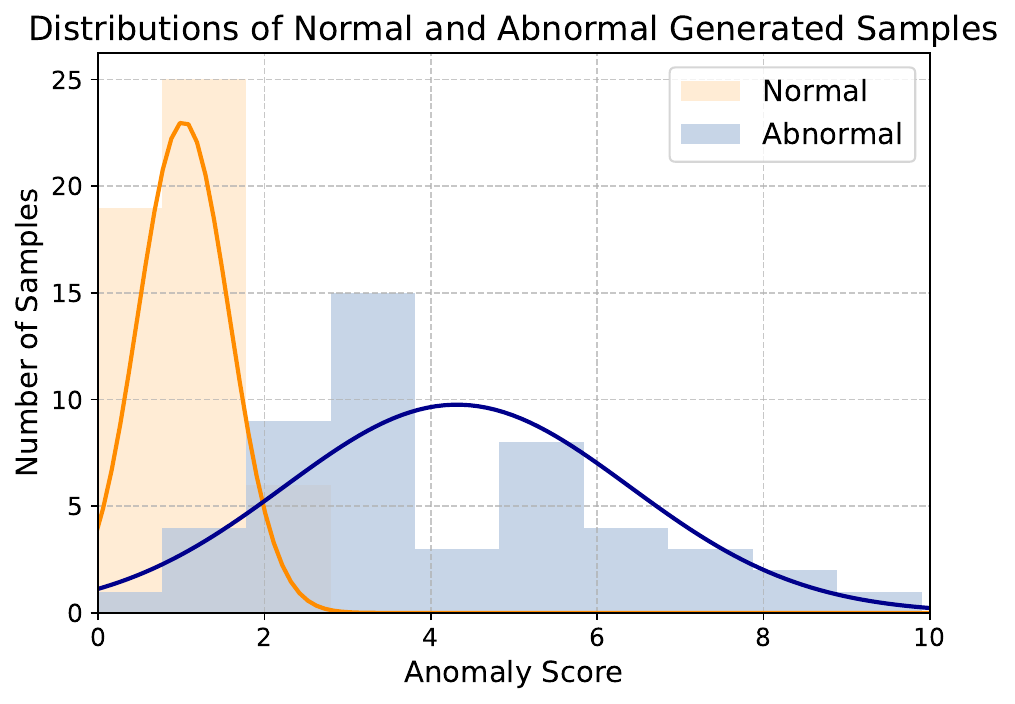} 
  \caption{Histograms of the anomaly scores for 50 future frames generated by \texttt{Diffusion} on the HR-STC dataset, for both cases of conditioning on normal and abnormal past frames.}
  \label{fig:histogram}
\end{figure}

In this analysis, we assess the performance of anomaly detection by altering the method of aggregation. Given the infinite variations in executing both normal and abnormal actions, we generate $M$ sets of future frames. For each set, we calculate an anomaly score. As discussed in the main paper, for the purpose of statistically aggregating these scores, we explore four strategies: taking the mean, the median, the maximum distance, and the minimum distance. In the mean and median approaches, we derive either the mean or the median of all $M$ scores and allocate this value to the respective frame to evaluate its anomaly level. Regarding the maximum and minimum distance selector approach, the highest and lowest anomaly score among all scores is assigned to the frame respectively. Comparison between these four methods is shown in \cref{tab:table_aggregation10}, with the minimum distance approach demonstrating superior performance across the board. The \b{suboptiomal}\s{under} performance of the maximum distance strategy further supports the idea that generated future samples that are conditioned on normal motions are as diverse as those that are conditioned on anomalous motions. This is due to the fact that if normal conditioned future sampled were not diverse, both the maximum and minimum distance strategies would have resulted in identical outcomes. \cref{fig:histogram} further demonstrates the effectiveness of our proposed GiCiSAD method in generating a diverse range of samples conditioned on both normal and abnormal frames. As can be seen, when the model is conditioned on normal past frames, the generated future frames are diverse yet close to the ground truth, with low anomaly scores. Conversely, when conditioned on abnormal past frames, the generated frames remain diverse but deviate significantly from the ground truth.

%%%%%%%%%%%%%%%%%%%%%5
\begin{table}[tb]
\centering

\scalebox{0.9}{
\begin{tabular}{lcc}
\toprule
                        \textbf{Aggregation Strategy} \hspace{1em} & \textbf{HR-Avenue} \hspace{1em} & \textbf{HR-STC} \\ \midrule
Mean     & 89.5        & 77.8     \\
Median     & 89.5        & 77.9     \\
Maximum Distance     & 88.2        & 77.3     \\
Minimum Distance          & \textbf{89.6}        & \textbf{78} \\ \bottomrule
\end{tabular}}
\caption{Comparison between different aggregation strategies for 50 generation of future frames, assessed through the AUROC metric on the HR-Avenue and HR-STC datasets.}
\label{tab:table_aggregation10}
\end{table}

%%%%%%%%%%%%%%%%%%%%555

\section{Baselines}
\label{sec:Baselines}

As mentioned in the main paper, we compare GiCiSAD against SOTA methods. Details of each method are described below.

\begin{enumerate}
    % \item Conv-AE \cite{hasan2016learning} learns the patterns of normal actions using autoencoder-based models. They develop two separate models: one uses handcrafted local features with a fully connected autoencoder, and the other employs a fully convolutional autoencoder for end-to-end learning. The reconstruction errors are used to detect anomalies.
    % %
    % \item Frame-Pred \cite{liu2018future} employs a U-Net model to generate future frame. They enhance the prediction of normal events by applying not only spatial constraints on intensity and gradient, but also introducing a motion constraint to ensure the consistency between predicted and actual frames. 
    % %
    % \item MPED-RNN \cite{morais2019learning} utilizes an innovative Message-Passing Encoder-Decoder Recurrent Network designed to analyze movements by separating them into global and local components. Within this architecture, a single encoder is employed, which branches into two distinct decoder heads. This dual-decoder setup leverages both reconstruction and prediction loss functions to effectively identify anomalies. 
    %
    \item GEPC \cite{markovitz2020graph} analyzes human poses through graphs. By mapping these graphs into a latent space and clustering them, they represent each action based on its soft-assignments to these clusters, akin to a "bag of words" model where actions are defined by their resemblance to foundational action-words. They then employ a Dirichlet process-based mixture model to classify actions as normal or anomalous.
    %
    % \item Multi-timescale Prediction \cite{rodrigues2020multi} adopts a fully-connected layered prediction method across various timescales to identify both short and long-term anomalies more effectively than single timescale methods. 
    %
    % \item Normal Graph \cite{luo2021normal} 
    % introduces a spatial-temporal graph convolutional network for skeleton-based video anomaly detection. They propose a predictive network where joints (nodes)
    % of abnormal events will be outliers of the graph.
    %
    \item PoseCVAE \cite{jain2021posecvae} predicts future pose trajectories based on a sequence of past normal poses, aiming to learn a conditional posterior distribution that characterizes the normal data, using a conditional variational autoencoder. They also propose a self-supervised component to enhance the encoder and decoder's ability to capture the latent space representations of human pose trajectories effectively. They imitate abnormal poses in the embeded space and use a binary cross-entropy loss along with the standard conditional variational autoencoder loss function.
    % \item BiPOCO \cite{kanu2022bipoco} proposes a Gated Recurrent Unit-based trajectory reconstructor that detects pedestrian anomalies in videos by reconstructing future frames using encoded past and future frames.
    %
    \item STGCN-LSTM \cite{li2022human} merges spatial-temporal graph convolutional autoencoder and Long Short-term Memory networks. They use  reconstruction and future prediction errors for detecting anomalies.
    \item COSKAD \cite{flaborea2023contracting} utilizes a graph convolutional network to encode skeletal human motions, and learns to project skeletal kinematic embeddings onto a latent hypersphere of minimal volume for video anomaly detection. COSKAD innovates by proposing three types of latent spaces: the traditional Euclidean, their proposed spherical and hyperbolic spaces. 
    \item MoCoDAD \cite{flaborea2023multimodal} utilizes autoencoder conditioned diffusion probabilistic models to generate a variety of future human poses. Their autoencoder-based approach conditions on individuals' past movements and leverages the enhanced mode coverage of diffusion processes to produce diverse yet plausible future motions. By statistically aggregating these potential futures, the model identifies anomalies when the forecasted set of motions diverges significantly from the observed future. 
    \item TrajREC \cite{stergiou2024holistic} 
    leverages multitask learning to encode temporally occluded trajectories, jointly learn latent representations of the occluded segments, and reconstruct trajectories based on expected motions across different temporal segments.

\end{enumerate}

\section{Weaker Forms of Conditioning Mechanism}
\label{sec:Conditioning_Mechanism}
\begin{figure}[tb]
  \centering
  \includegraphics[width=1.02\linewidth]{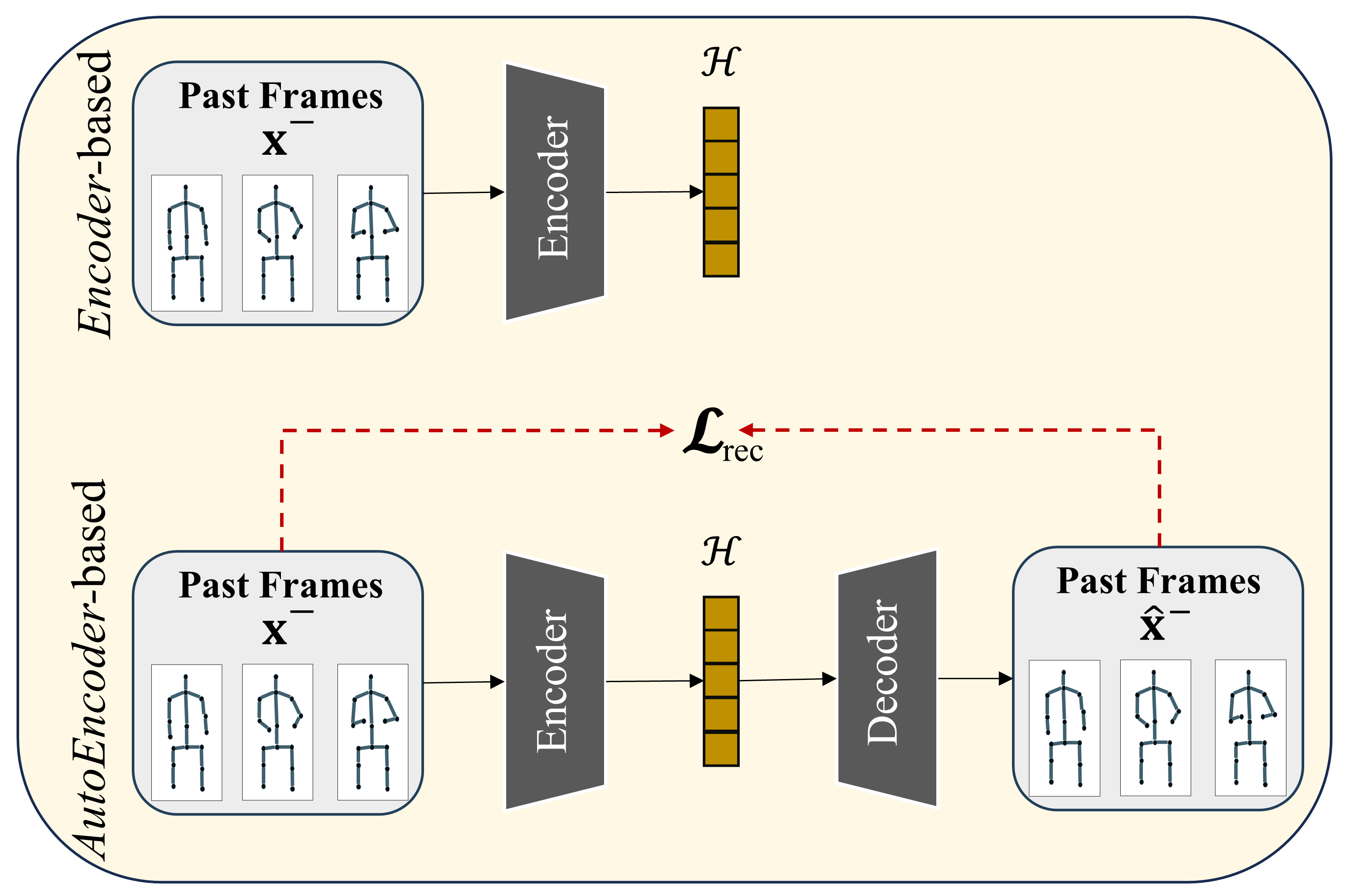} 
  \caption{Comparison of \textit{Encoder}-based and \textit{Autoencoder}-based conditioning mechanisms.}
  \label{fig:AE-E}
\end{figure}

This section elaborates on the \textit{Encoder}-based and \textit{AutoEncoder}-based conditioning mechanisms \cite{flaborea2023multimodal} that are used for comparison with our proposed \textit{Graph}-based approach, mentioned in the ablation study of the main paper. The objective of conditioning mechanism is to generate an efficient latent representation of past frames, 
$\mathcal{H}$, to effectively guide the \texttt{Diffusion} process. The architecture of these two conditioning mechanisms is illustrated in \cref{fig:AE-E}. $\mathcal{H}$ will be used as the conditioning signal to guide the \texttt{Diffusion}, where the architecture of \texttt{Diffusion} remains unchanged. The \textit{Encoder}-based method introduces no additional loss to the model. Conversely, the \textit{AutoEncoder}-based approach incorporates the reconstruction loss of the past frames into the \texttt{Diffusion} loss, thereby modifying the overall loss calculation as follows.

\begin{equation}
    \mathcal{L} = \lambda \mathcal{L}_\text{rec} +  \mathcal{L}_{\text{diffusion}},
\end{equation}
where $\lambda$ is $0.1$. From an architectural perspective, the encoder features a channel sequence of $(32, 16, 32)$, incorporating a bottleneck dimension of $32$ and a latent projector with a dimensionality of $16$. 

%%%%%%%%%%%%%%%%%%%%555
%%%%%%%%%%%%%%%%%%%%%%%%%%%%%%%%%%555

\section{Visualization of \textit{Intra}-Community Shuffling Approach}
\label{sec:Jigsaw_Puzzles_type}

\begin{figure*}[t]
  \centering
  \includegraphics[width=0.7\textwidth]{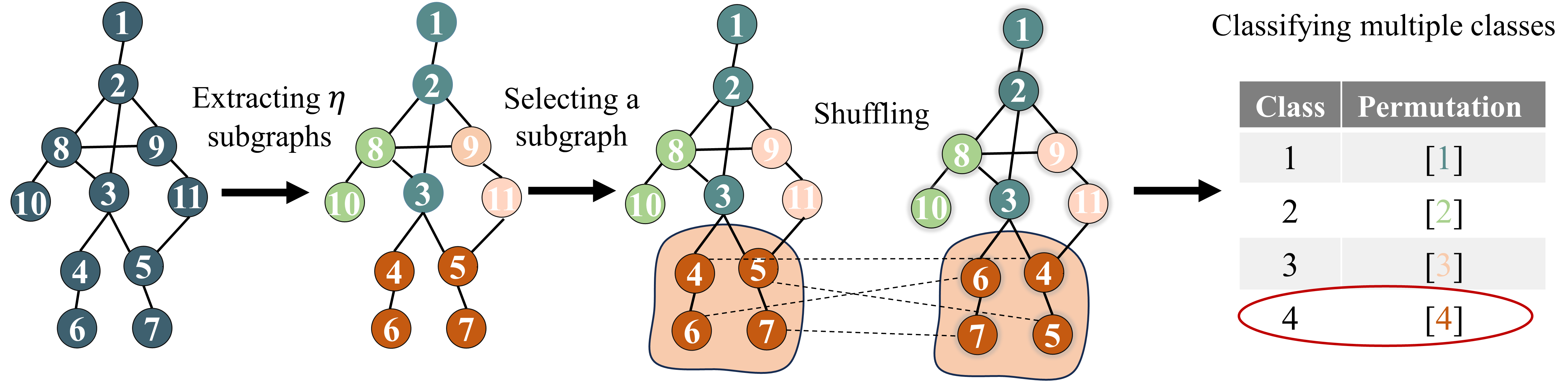}

   \caption{Visualization of the \textit{Intra}-Community shuffling approach. Nodes with the same color formulate a subgraph. Note that although each node is required to have $\delta$ connections, for improved visualization, this property is not strictly maintained in the figure.}
   \label{fig:diverce}
\end{figure*}

The visualization of the \textit{Intra}-Community shuffling approach is shown in \cref{fig:diverce}. A detailed description of this approach has previously been provided in the ablation study section, "Types of Graph-based Jigsaw Puzzles," of the main paper.

%%%%%%%%%%%%%%%%%%%%555
% \begin{figure}[tb]
%   \centering
%   \includegraphics[width=0.9\textwidth]{images/intra_community.png} 
%   \caption{Visualization of the \textit{Intra}-Community shuffling approach. Nodes with the same color formulate a subgraph. Note that although each node is required to have $\delta$ connections, for improved visualization, this property is not strictly maintained in the figure.}
%   \label{fig:diverce}
% \end{figure}

%%%%%%%%%%%%%%%%%%%%%%%%%%%%55
\begin{figure*}[t]
  \centering
  \includegraphics[width=0.7\textwidth]{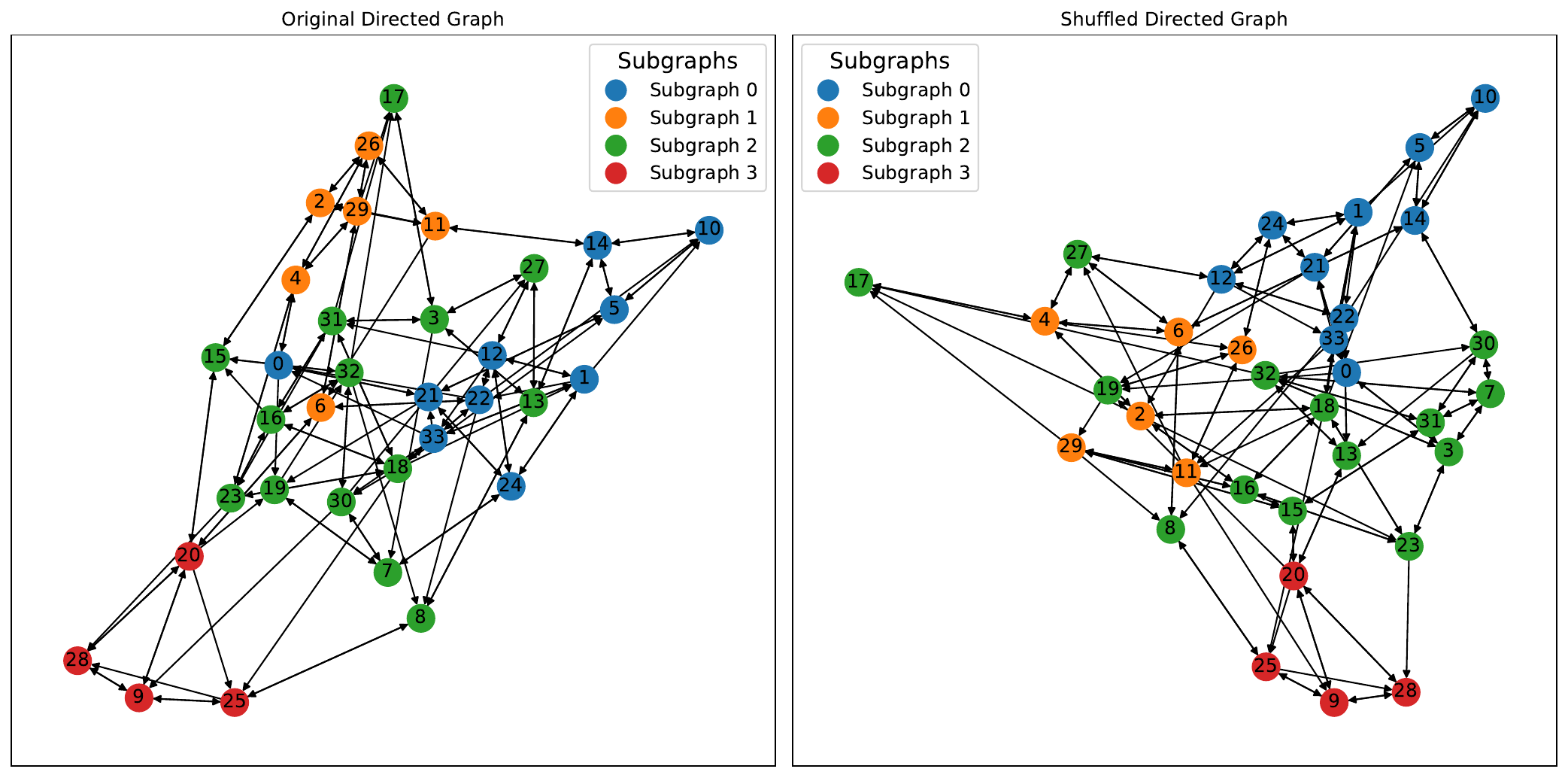}

   \caption{\textit{Inter}-Community shuffling process between Subgraph 2 and Subgraph 1. }
   \label{fig:shuffle}
\end{figure*}

\section{Visualization of \textit{Inter}-Community Shuffling Approach on Real Constructed Graphs}
\label{sec:visualization}

While we have presented a simple and easy-to-understand visualization of our \textit{Inter}-community shuffling approach in Fig. 2 of the main paper, we provide \cref{fig:shuffle} for a more detailed view of the \textit{Inter}-community shuffling process on the real constructed graphs. Graphs include 34 nodes (joints), $\delta$ and $\eta$ are set to 4. Note that the graphs (before and after shuffling) are directed, indicating that connections are not inherently symmetric. In this figure, Subgraph 2 (depicted in green) is shuffled with Subgraph 1 (depicted in orange). Specifically, the densest nodes of Subgraph 2, namely, $\{32, 3, 31, 30, 7, 13\}$, are shuffled with nodes $\{29, 4, 6, 11, 26, 2\}$ from Subgraph 1, respectively. After the shuffling process, while nodes of the smaller subgraph, i.e., Subgraph 1, stay connected, the intra-connections of the larger subgraph, i.e., Subgraph 2, undergoes significant changes, nearly dividing it into two distinct parts. It should be noted that the other two subgraphs, i.e., Subgraphs 0 and 3, retain their connections, while only their spatial positioning is changed for better visualization in the figure.

\end{document}